\documentclass[letterpaper, 10 pt, conference]{ieeeconf}  

\IEEEoverridecommandlockouts                              

\newcommand{\etal}{{\,\textit{et al.}}}

\newcommand*\widebar[1]{%
   \hbox{%
     \vbox{%
       \hrule height 0.5pt 
       \kern0.3ex
       \hbox{%
         \kern-0.1em
         \ensuremath{#1}%
         \kern-0.1em
       }%
     }%
   }%
}

\title{\LARGE \bf
Self-Supervised Person Detection in 2D Range Data\\using a Calibrated Camera}

\author{Dan Jia$^{1}$, Mats Steinweg$^{1}$, Alexander Hermans$^{1}$, and Bastian Leibe$^{1}$
\thanks{$^{1}$All authors are with the Visual Computing Institute, RWTH Aachen University.
        {\tt\footnotesize \{jia,hermans,leibe\}@vision.rwth-aachen.de, mats.steinweg@rwth-aachen.de}}%
}

\usepackage{silence}
\usepackage[dvipsnames]{xcolor}
\usepackage{graphicx}
\usepackage{amssymb}
\usepackage{amsmath}
\usepackage{multirow}
\usepackage{colortbl}
\usepackage{color}
\usepackage{graphicx}
\usepackage{hyperref}
\usepackage{xspace}
\usepackage{tabularx}
\usepackage{tikz}
\usetikzlibrary{calc,shapes,arrows,backgrounds}
\usepackage{pgfplots}
\pgfplotsset{compat=1.16}
\usepackage{pgfplotstable}
\usepackage{siunitx}
\usepackage{booktabs}
\usepackage{csvsimple}
\usepackage{makecell}
\usepackage{eqnarray}
\usepackage{bbm}
\usepackage{textcomp}
\usepackage{gensymb}
\usepackage{pifont}
\usepackage{subcaption}
\usepackage{dsfont}
\usepackage{tikzscale}
\usepackage{balance}
\usepackage{hyperref}

\definecolor{new_red}{HTML}{E24A33}
\definecolor{new_blue}{HTML}{348ABD}
\definecolor{new_purple}{HTML}{D175F0}
\definecolor{new_green}{HTML}{8EBA42}
\definecolor{new_orange}{HTML}{F9A91F}
\definecolor{new_gray}{HTML}{777777}
\definecolor{new_pink}{HTML}{FF69B4}
\definecolor{new_brown}{HTML}{8B4513}
\definecolor{new_blue_dark}{HTML}{1269B0}
\definecolor{new_red_dark}{HTML}{BF534F}
\definecolor{new_green_dark}{HTML}{1e942e}
\definecolor{new_yellow_dark}{HTML}{c99436}
\definecolor{new_cyan_dark}{HTML}{3ab7c7}
\definecolor{new_orange_dark}{HTML}{f05d02}

\pgfplotsset{
    precrec/.style={
        inner sep=0pt,outer sep=0pt,
        ylabel style={font=\scriptsize,yshift=-18pt},
        xlabel style={font=\scriptsize,yshift=6pt},
        width={1.1\linewidth},
        height={1.1\linewidth},
        yticklabel style = {font=\scriptsize,xshift=-0.3ex},
        xticklabel style = {font=\scriptsize,yshift=-0.3ex},
        legend image post style={line width =1.5pt},
        tick label style={/pgf/number format/assume math mode=true},
        tick align=outside,
        major tick length=2pt,
        every tick/.style={black, thin},
        tick pos=left,
        axis line style={draw=none},
        xtick={0,20,...,100},
        ytick={0,20,...,100},
        xlabel={Recall [\%]},
        ylabel={Precision [\%]},
        xmin=-2,
        xmax=102,
        ymin=-2,
        ymax=102,
        axis background/.style={fill=black!11!white},
        grid=both,
        grid style={white}
    }
}

\pgfplotsset{legend image code/.code={%
                \draw[mark repeat=2,mark phase=2]
                plot coordinates {
                (0cm,0cm)
                (0.3cm,0cm)        
                (0.6cm,0cm)         
                };%
            }
}

\def\addlegendimage{\csname pgfplots@addlegendimage\endcsname}

\makeatletter
\DeclareRobustCommand\onedot{\futurelet\@let@token\@onedot}
\def\@onedot{\ifx\@let@token.\else.\null\fi\xspace}

\def\etal{\emph{et al}\onedot}
\makeatother

\newcolumntype{Y}{>{\centering\arraybackslash}X}

\begin{document}

\maketitle
\thispagestyle{empty}
\pagestyle{empty}



\begin{abstract}
Deep learning is the essential building block of state-of-the-art person detectors in 2D range data.
However, only a few annotated datasets are available for training and testing these deep networks, potentially limiting their performance when deployed in new environments or with different LiDAR models.
We propose a method, which uses bounding boxes from an image-based detector (\textit{e.g.} Faster R-CNN) on a calibrated camera to automatically generate training labels (called \textit{pseudo-labels}) for 2D LiDAR-based person detectors.
Through experiments on the JackRabbot dataset with two detector models, DROW3 and DR-SPAAM, we show that self-supervised detectors, trained or fine-tuned with pseudo-labels, outperform detectors trained only on a different dataset.
Combined with robust training techniques, the self-supervised detectors reach a performance close to the ones trained using manual annotations of the target dataset.
Our method is an effective way to improve person detectors during deployment without any additional labeling effort, and we release our source code to support relevant robotic applications.
\end{abstract}

\section{INTRODUCTION}
\label{sec:introduction}

2D LiDARs provide accurate range measurements with a large field of view, often greater than 200 degrees, at an affordable price, and are a popular sensor choice for many robotic tasks, including person detection.
While early approaches for detecting persons in 2D range data focused on heuristics with hand-crafted features \cite{Leigh15ICRA,Arras07ICRA},
recent studies used convolutional neural networks and further improved the detection results \cite{Beyer18RAL,Jia20arXiv}.

\begin{figure}[ht]
    \centering
 \includegraphics[width=\linewidth]{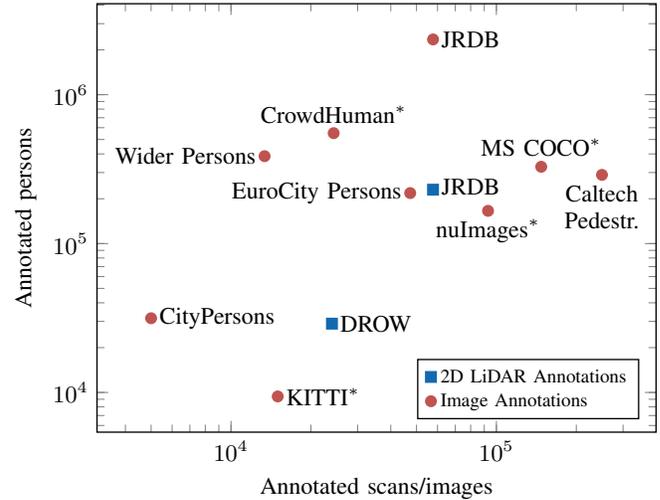}
    \caption{
    Only few 2D LiDAR datasets with person annotations exist, while a vast amount of image-based annotations are available.
    We utilize the image-based annotations to train 2D LiDAR-based person detectors by generating pseudo-labels from image-based detections.
    $^*$: Numbers are estimated based on person-per-image statistics of the training set.
    }
    \label{fig:datasets}
\end{figure}

Deep learning has become an integral part of modern detection algorithms, whether from 2D range data \cite{Beyer18RAL,Jia20arXiv} or images \cite{Tan20CVPR,He17ICCV,Redmon16CVPR,Liu16ECCV,Ren15NIPS}. 
The success of these detectors hinges upon the availability of large and high-quality datasets.
Over the past years, significant effort has gone into labeling images with bounding boxes or segmentation masks, whereas relatively little attention has been given to annotating 2D range data (see~Fig.\,\ref{fig:datasets}).
The few available datasets for 2D LiDARs do not possess enough diversity in terms of the surrounding environments and sensor models.
Networks trained solely on these data may not generalize well at deployment, where both the environment and the sensor specification are likely to differ from those encountered during training.

To overcome the limitation imposed by insufficient training data,
we propose a method to automatically generate labels for training LiDAR-based person detectors using the output of an image-based detector on a calibrated camera.
Given a person bounding box in an image, our method takes the 2D LiDAR points that fall within the box frustum and uses a clustering algorithm to locate the person in the LiDAR coordinate.
The estimated locations of persons in the scene, as well as a set of negative points, are used as pseudo-labels for training a detector.
This allows a robot to train or fine-tine a laser-based detector in a self-supervised fashion.
We empirically demonstrate the validity of the pseudo-labels and show that they can be used for both training and fine-tuning a 2D LiDAR-based person detector.
Additionally, we experiment with robust training techniques to further improve the detector performance. 
Our method is an effective way to boost the performance of a person detector during deployment without any additional labeling effort, and has great potential for many robotic applications. 
In summary, the main contributions of this work are:
\begin{itemize}
    \item We propose a method to automatically generate pseudo-labels for training 2D LiDAR-based person detectors, leveraging the output of an image-based detector and known extrinsic camera calibration.
    \item We demonstrate that the generated pseudo-labels can be used to train or fine-tune a person detector and experiment with robust training techniques to further improve its performance.     
    \item We release our code, implemented in PyTorch with an easy-to-use detector ROS node, for robotic applications.\footnote{\url{https://github.com/VisualComputingInstitute/2D_lidar_person_detection}} 
\end{itemize}


\section{RELATED WORK}
\label{sec:related_work}

\subsection{LiDAR-based Person Detection}
Person detection from 2D range data has a long-standing history in the robotics community.
Early approaches~\cite{Fod02ICRA,Scheutz04IROS,Schulz03IJRR} focus on tracking moving blobs in sequential LiDAR scans.
These blobs are detected using manually engineered heuristics in a non-learning fashion.
Later developments~\cite{Leigh15ICRA,Arras07ICRA,Pantofaru10ROS} improved the detection stage by using supervised learning techniques.
In these approaches, a clustering algorithm is first applied over scan points, using clues like proximity~\cite{Leigh15ICRA} or jump distance~\cite{Arras07ICRA}.
A set of hand-crafted features is then extracted for each cluster, and these features are used to train a simple classifier~(\textit{e.g.}~AdaBoost).
In the case of~\cite{Leigh15ICRA,Pantofaru10ROS}, additional heuristics from tracking are employed to post-process the detected clusters.  

Most recent developments~\cite{Beyer16RAL,Beyer18RAL,Jia20arXiv} use deep learning techniques to detect persons directly from data, without manually engineered heuristics or features.
The DROW detector~\cite{Beyer16RAL} is the first deep learning-based walking aid detector working on 2D range data and was later extended to additionally detect persons~\cite{Beyer18RAL}.
The current state-of-the-art method is the DR-SPAAM detector~\cite{Jia20arXiv}, which leverages a temporal aggregation paradigm to incorporate multiple scans into the detection process, alleviating the problems associated with the low information content in a single LiDAR scan while retaining real-time computation on a mobile robotic platform.

\subsection{Automatic Label Generation}
Supervised learning approaches demand intense effort to manually collect and annotate data for the purpose of training and testing.
Automatic label generation has been attempted to reduce the required labeling effort.
For example, Leigh \etal \cite{Leigh15ICRA} generate positive training examples by positioning a 2D LiDAR in an open environment populated with people and negative examples by moving a LiDAR in an environment devoid of people.
This method limits the training examples to a few simple scenarios and cannot be adapted for dynamic data collection at deployment time.
Aguirre \etal \cite{Aguirre2019IJCIS} use Mask R-CNN~\cite{He17ICCV} with a calibrated RGB-D camera to automatically label 2D LiDAR scans for person detection.
However, no evaluation of the labels' accuracy is performed.
Furthermore, only very simple LiDAR-based person detectors are used and both the training and the evaluation are conducted using (potentially incorrect) generated labels, with no evaluation using real annotated data.
The method also relies on depth measurements from an RGB-D camera, imposing an additional sensor requirement. 
Automatic label generation has also been attempted for 3D LiDARs.
Piewak \etal \cite{Piewak18ECCV} and Wang \etal \cite{Wang19RAL} train point-cloud segmentation networks, using segmentation results on matching pixels as supervision.

In this work, we propose to automatically generate training data for a 2D LiDAR-based person detector using a calibrated RGB camera and image-based person detections.
Compared to~\cite{Leigh15ICRA}, our pseudo-labels are dynamically generated and do not rely on specific conditions of the environment.
Unlike~\cite{Aguirre2019IJCIS}, our method operates on normal RGB cameras and does not require expensive segmentation on images or pixel-precise calibration between sensors.
We conduct extensive experiments to analyze the quality of pseudo-labels and empirically prove their value for training state-of-the-art person detectors.

\subsection{Learning with Noisy Labels}

Training neural networks with imperfect datasets is becoming an increasingly relevant topic.
Proposed methods range from robust loss functions~\cite{Ghosh17AAAI,Zhang18NIPS,Wang2019ICCV,Menon2020ICLR}, or noise modeling~\cite{xiao2015learning,Goldberger17ICLR,Han18NIPSMasking}, to sample selection~\cite{Pawan10NIPS,Jiang18ICML,Han18NIPS}, or re-weighting samples~\cite{Ren18ICML,Shu19NIPS}.
Regularization techniques~\cite{Srivastava14JMLR,Jindal16ICDM,Menon2020ICLR} have also been shown to reduce the effect of label noise.
For a more thorough study on this topic, we refer readers to surveys~\cite{Zhang16AIR,Algan19arXiv,Song20arXiv}.

In our work, we experiment with the~\textit{partially Huberized cross-entropy loss}~\cite{Menon2020ICLR} and the~\textit{mixup} regularization~\cite{Zhang2018ICLR} to deal with the inherent noise of pseudo-labels.
These two methods were picked for their applicability\textemdash they do not rely on specific noise properties, nor do they impose additional constraints (\textit{e.g.} a small set of clean training data).


\section{Generating Pseudo-Labels}
\label{sec:generating_pseudo_labels}

We use a calibrated camera to generate pseudo-labels for training a 2D LiDAR-based person detector.
These pseudo-labels include the location of persons in the LiDAR coordinate frame $\{(p_{x, i}, p_{y,i})\}_i$, and a set of scan points that belong to the background of the scene.

We first use an object detector (\textit{e.g.}~Faster~R-CNN~\cite{Ren15NIPS}) to obtain person bounding boxes.
From all bounding boxes, a subset is selected using the following constraints:
\begin{itemize}
    \item \textit{classification score} greater than a threshold $T_{c}$,
    \item \textit{aspect ratio}, the ratio between width and height, smaller than $T_{AR}$, 
    \item \textit{overlap ratio}, with any other bounding box is smaller than $T_o$. The overlap ratio is defined as the intersection area divided by the area of the box.
\end{itemize}
The goal is to select boxes from which the location of persons can be confidently extracted, rather than locating \textit{all} persons in the scene (\textit{i.e.}~we favor precision over recall in this step).

Given a selected bounding box, we estimate the center location $(p_{x,i}, p_{y,i})$ of the person in the LiDAR coordinate frame.
Utilizing the known camera calibration, we project the LiDAR points onto the image and extract points that fall within the bottom half of the bounding box (since the LiDAR is mounted at the height of the lower body).
These points either correspond to a person or to the background (see~Fig.\,\ref{fig:qualitative_results}).
To localize the person, we first run a $k$-means clustering in the range space with $k=2$, which groups points into a close and a far cluster.
We take the average 2D location of points in the close cluster as the initial estimation, and iteratively refine this estimation using a mean shift procedure with a circular kernel of 0.5~$m$ radius.
The mean shift result is used as the estimated person location.
This proposed method assumes that the person belongs to the foreground of the scene and is the dominant object in the cropped LiDAR scan, which is typically satisfied by the content of detection bounding boxes.

LiDAR points that do not project to any bounding box (including discarded boxes) are taken as the negative training samples.
For increased robustness, we enlarge the width of each bounding box by a factor of 0.1 when generating the negative samples.


\section{Person Detection with Pseudo-Labels}

\subsection{DROW3 and DR-SPAAM Detector}

We experiment with two state-of-the-art person detectors, DROW3~\cite{Beyer18RAL} and its successor DR-SPAAM~\cite{Jia20arXiv}.
The DROW3 detector takes as input a 2D LiDAR scan, expressed as a one-dimensional vector of range measurements.
For each point, it outputs a classification label and, for the positive points, a location offset to the person center.
This is accomplished by pooling a small window of neighboring points, which are processed by a 1D convolutional neural network.
DR-SPAAM improves upon this approach by introducing a spatial attention and auto-regressive model that integrates temporal information to improve detection performance.
Thus, it requires the input to be a sequence of scans.
We refer readers to~\cite{Beyer18RAL,Jia20arXiv} for more details.

We use the generated pseudo-labels to train DROW3 and DR-SPAAM.
For supervising the classification branch, we use points less than 0.4~$m$ away from an estimated person center $(p_{x}, p_{y})$ as the positive samples, and the marked out background points as the negative samples.
For supervising the regression branch, we use points less than 0.8~$m$ away from an estimated person center.
To increase robustness, we discard pseudo-labels with less than five surrounding positive points.
Points that are neither close to a person nor marked as a background are ignored during training.


\subsection{Robust Training}

In the default setup, the classification branch of DROW3 and DR-SPAAM is supervised with the cross-entropy loss, which is prone to label noise in the training samples~\cite{Ghosh17AAAI,Wang2019ICCV}.
To limit the influence of wrongly generated pseudo-labels, we resort to robust training techniques.
We experiment with the~\textit{partially Huberized cross-entropy loss}~\cite{Menon2020ICLR}, a more robust loss function, and the~\textit{mixup} regularization~\cite{Zhang2018ICLR}.

\subsubsection{Partially Huberized cross-entropy loss}

The softmax cross-entropy loss is composed of a base loss (cross-entropy) with a sigmoid link.
Menon \etal \cite{Menon2020ICLR} introduce a composite loss-based gradient clipping, linearizing the base loss beyond a threshold while leaving the sigmoid link untouched.
The overall loss function takes the form:
\begin{equation}
    l = 
        \begin{cases}
            - \tau \cdot p + log(\tau) + 1, & \text{if}\ p\leq\frac{1}{\tau} \\
            - log(p), & \text{else,}
        \end{cases}
\end{equation}
which asymptotically saturates for $p\leq\frac{1}{\tau}$, and was proven to be robust against label noise.
The parameter $\tau$ should be set in proportion with the amount of noise in the training samples. 
In our experiments, we use $\tau=5$.

\subsubsection{Mixup regularization}

Given two training samples $(x_i,y_i)$ and $(x_j,y_j)$, Zhang \etal \cite{Zhang2018ICLR} proposed to construct augmented training samples:
\begin{align*}
    \tilde{x}&=\lambda x_i + (1-\lambda)x_j \\
    \tilde{y}&=\lambda y_i + (1-\lambda)y_j,
\end{align*}
with $\lambda\,{\sim}\,$Beta$(\alpha,\alpha)$ and the parameter $\alpha \in (0,\infty)$ controlling the augmentation strength. 
Unlike the conventional data augmentation which operates on a single sample (e.g. randomly flipping an image), \textit{mixup} generates virtual samples across training data.
It encourages linear behavior in-between training examples, and was shown to improve generalization of a network and increase its robustness against label noise.

To adapt \textit{mixup} for training a detection network, which includes both classification and regression, we use the following multi-task loss
\begin{equation}
    l_{total} = l_{reg} + (1 - w) \cdot l_{cls} + w \cdot l_{mixup},
\end{equation}
where $l_{reg}$ and $l_{cls}$ are the regression and classification loss without \textit{mixup} regularization, $l_{mixup}$ is the classification loss with \textit{mixup}, and $w$ is a weighting factor.
To avoid additional memory overhead, we split the cost into
\begin{align*}
    l_1 &= l_{reg} + (1 - w) \cdot l_{cls}\\
    l_2 &= w \cdot l_{mixup}
\end{align*}
and perform gradient descent over $l_1$ and $l_2$ sequentially at each training iteration.
In our experiment, we use $w=0.7$ and $\alpha=0.2$.


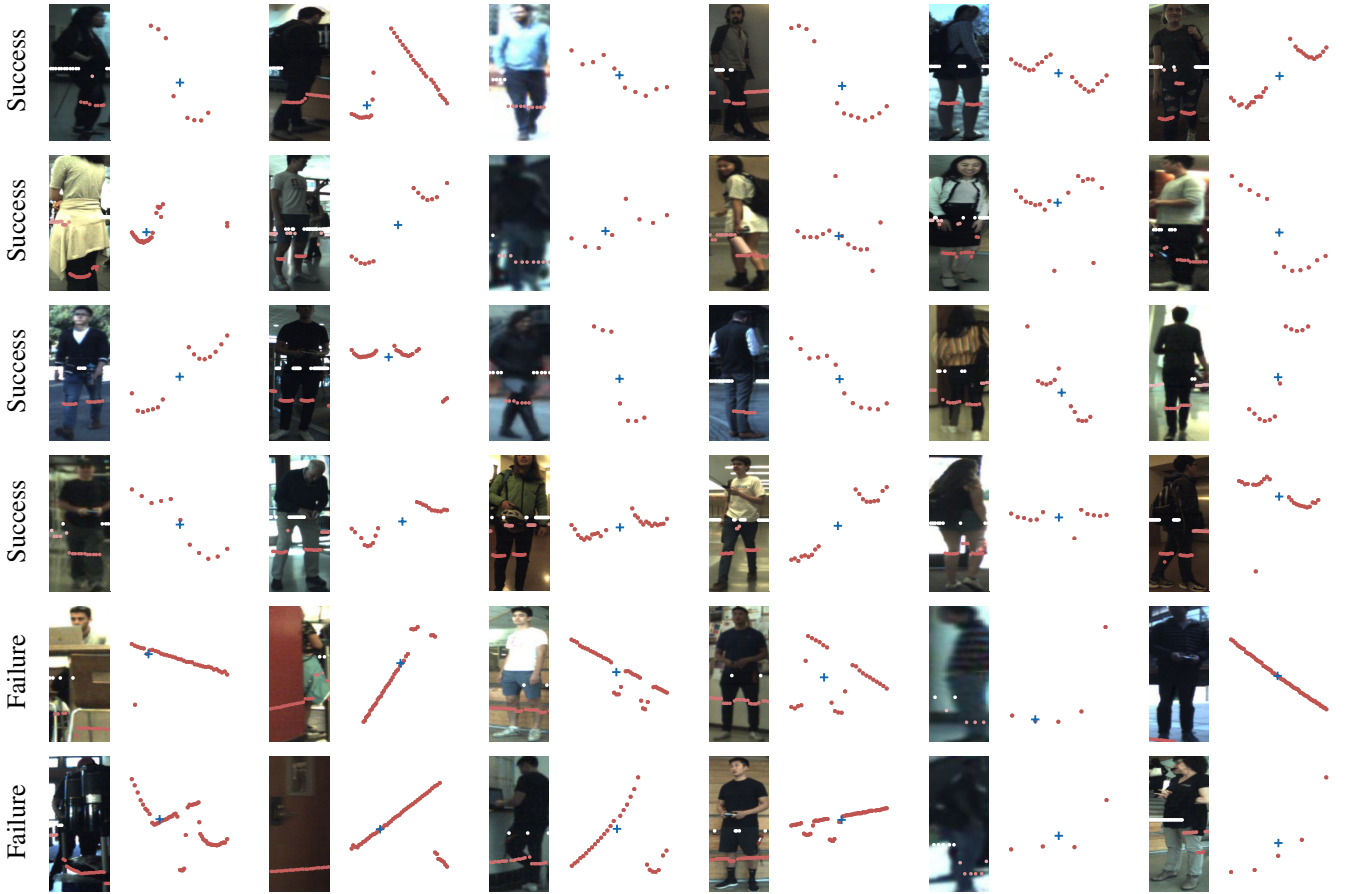
\begin{figure*}
    \newlength{\imw}
    \setlength{\imw}{0.8cm}
    \newlength{\imh}
    \setlength{\imh}{1.8cm}
    
    \centering
    
    \newcommand{\impairall}[1]{%
        \begin{subfigure}{\imw}%
            \includegraphics[width=\imw,height=\imh]{pics/pl_example/#1_im.pdf}%
        \end{subfigure}%
        \begin{subfigure}{\imh}%
            \includegraphics[width=\imh,height=\imh]{pics/pl_example/#1_pt.pdf}%
        \end{subfigure}%
    }%
    \newcommand{\impair}[1]{\impairall{#1}\hfill}
    \newcommand{\impairend}[1]{\impairall{#1}}

    \rotatebox[origin=c]{90}{Success}\hspace{2mm}
    \impair{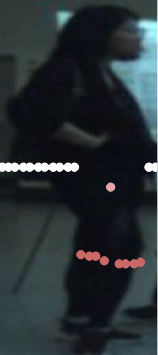}%
    \impair{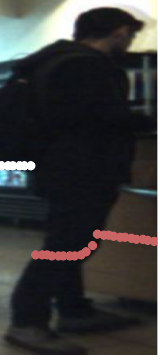}%
    \impair{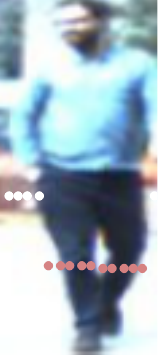}%
    \impair{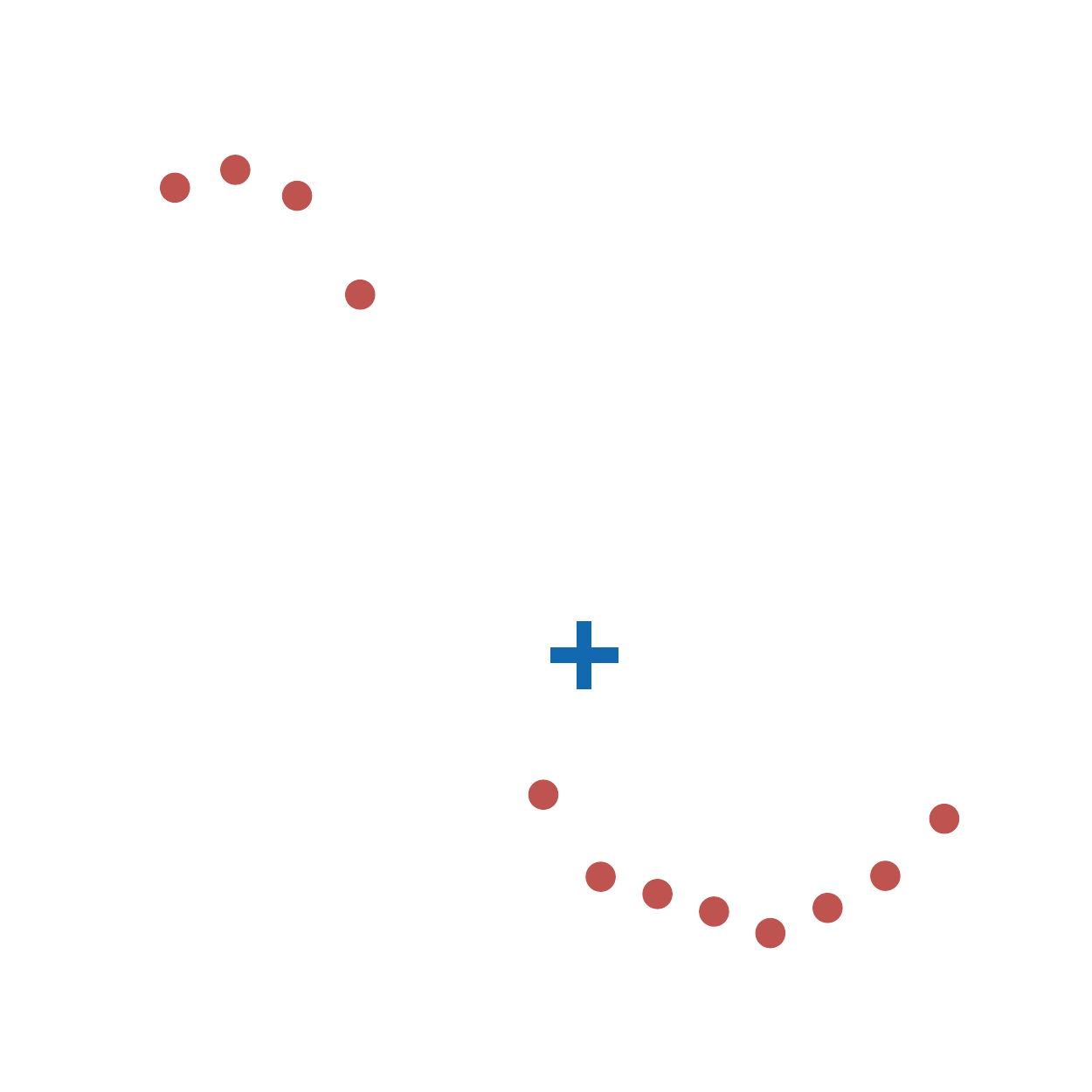}%
    \impair{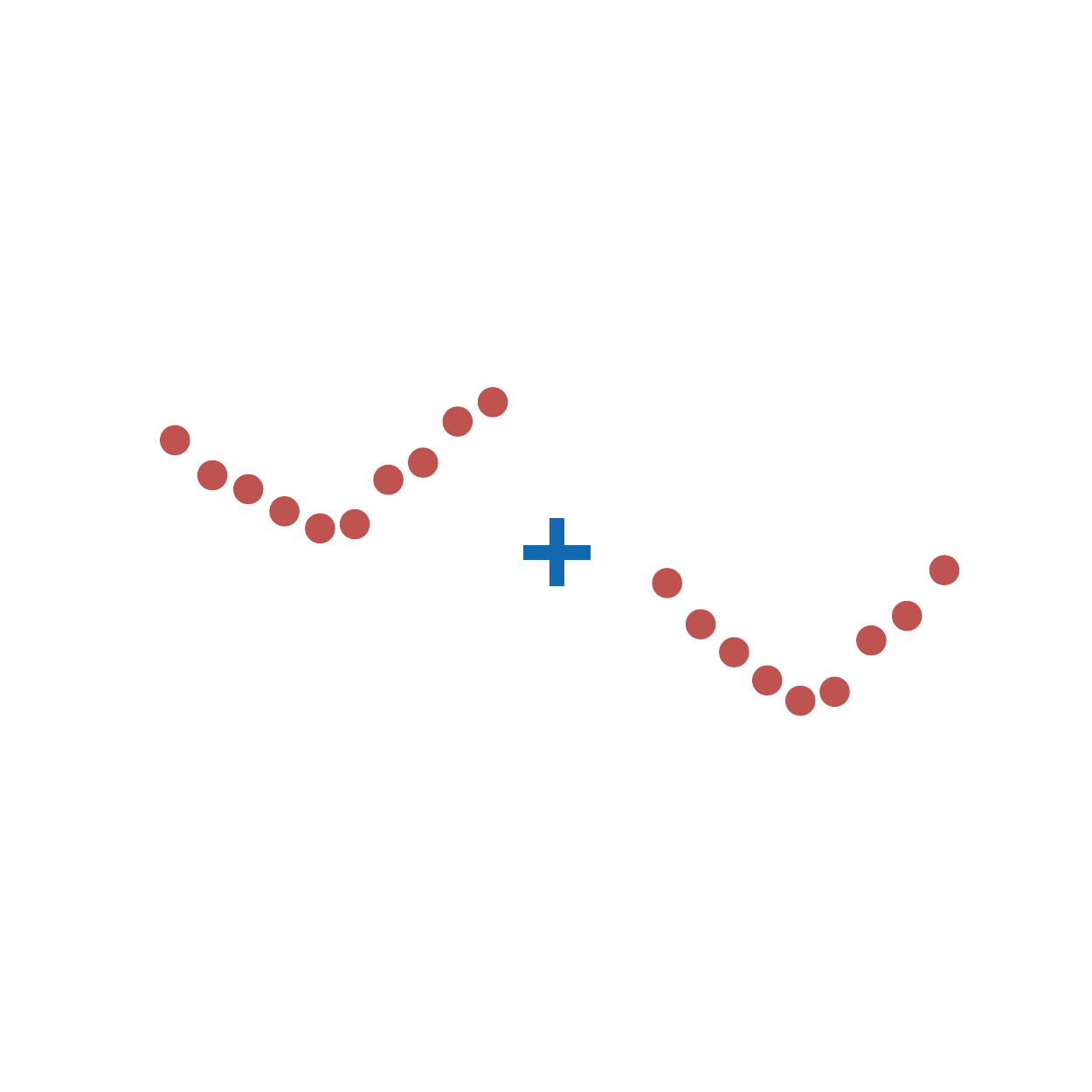}%
    \impairend{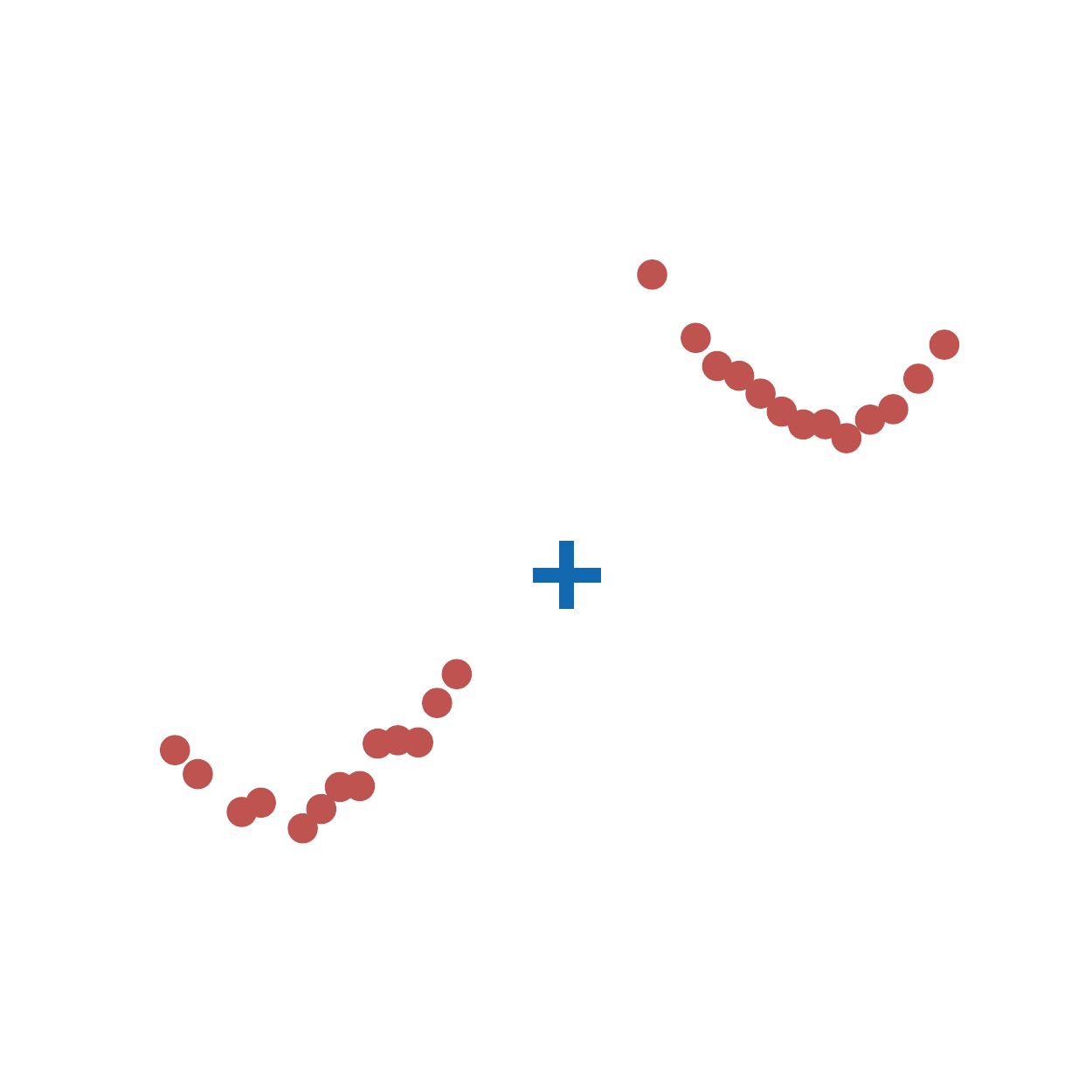}%
    \vspace{2mm}%
    
    \rotatebox[origin=c]{90}{Success}\hspace{2mm}
    \impair{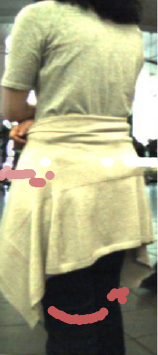}%
    \impair{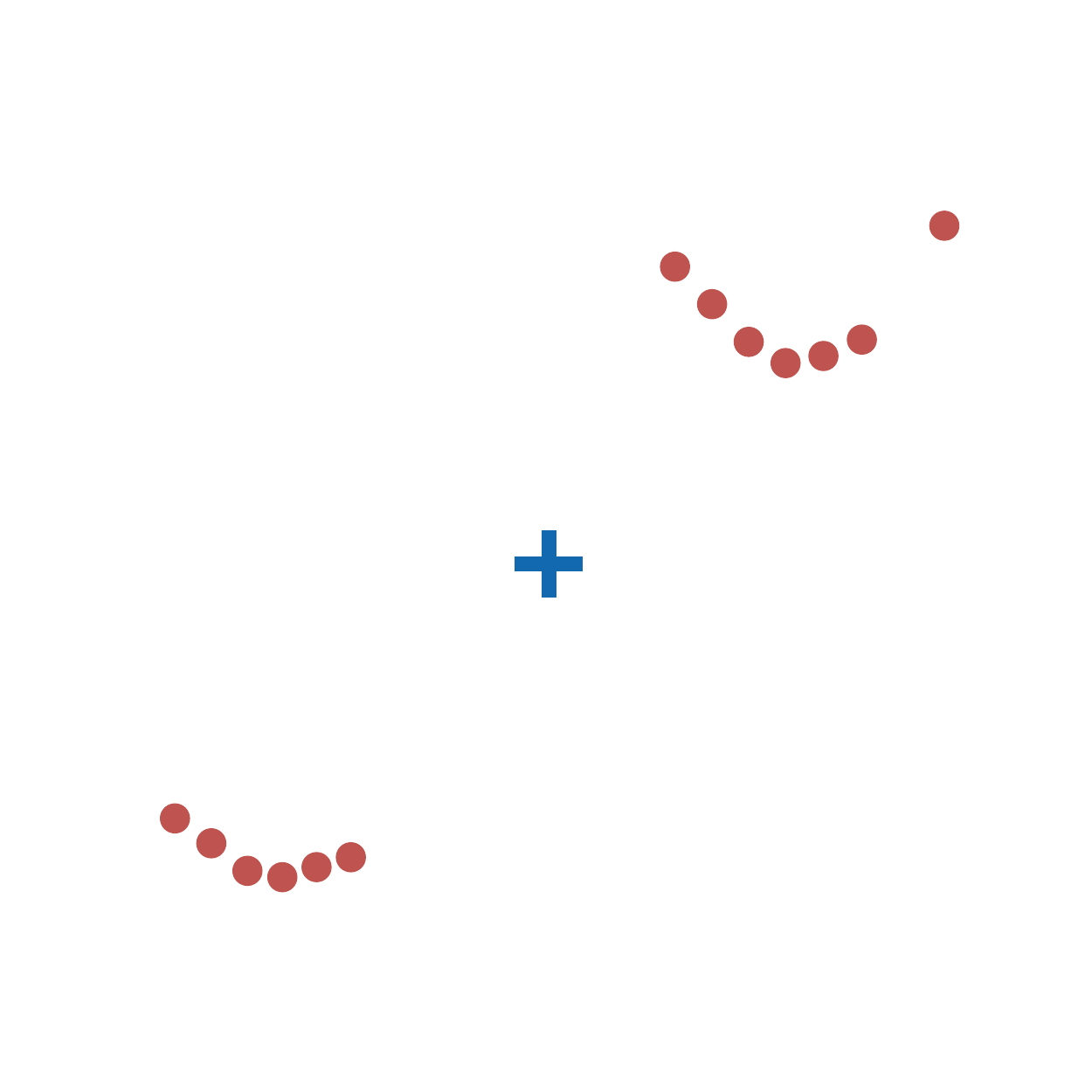}%
    \impair{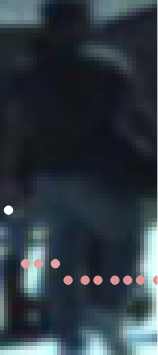}%
    \impair{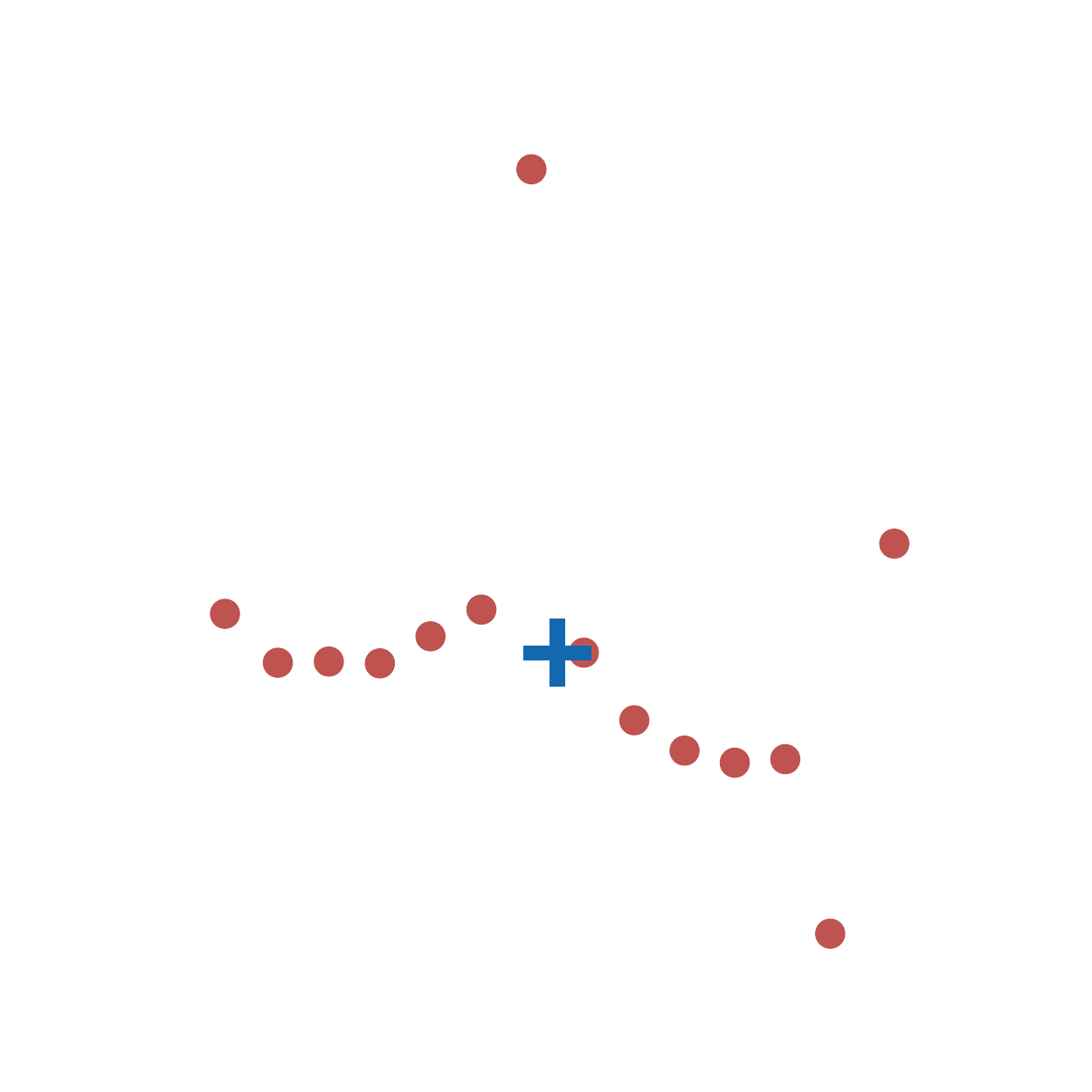}%
    \impair{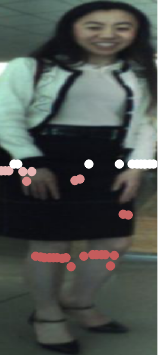}%
    \impairend{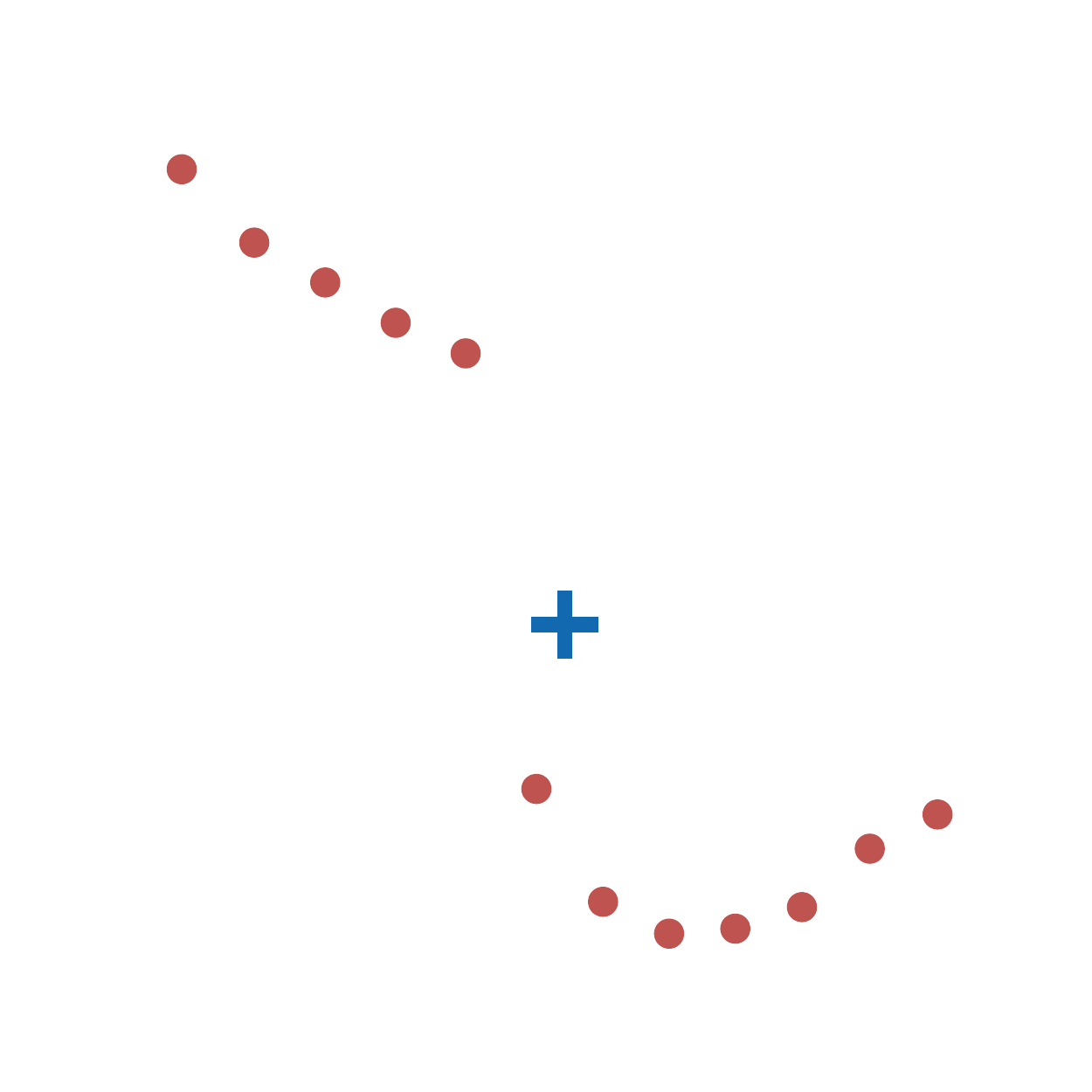}%
    \vspace{2mm}%
    
    \rotatebox[origin=c]{90}{Success}\hspace{2mm}
    \impair{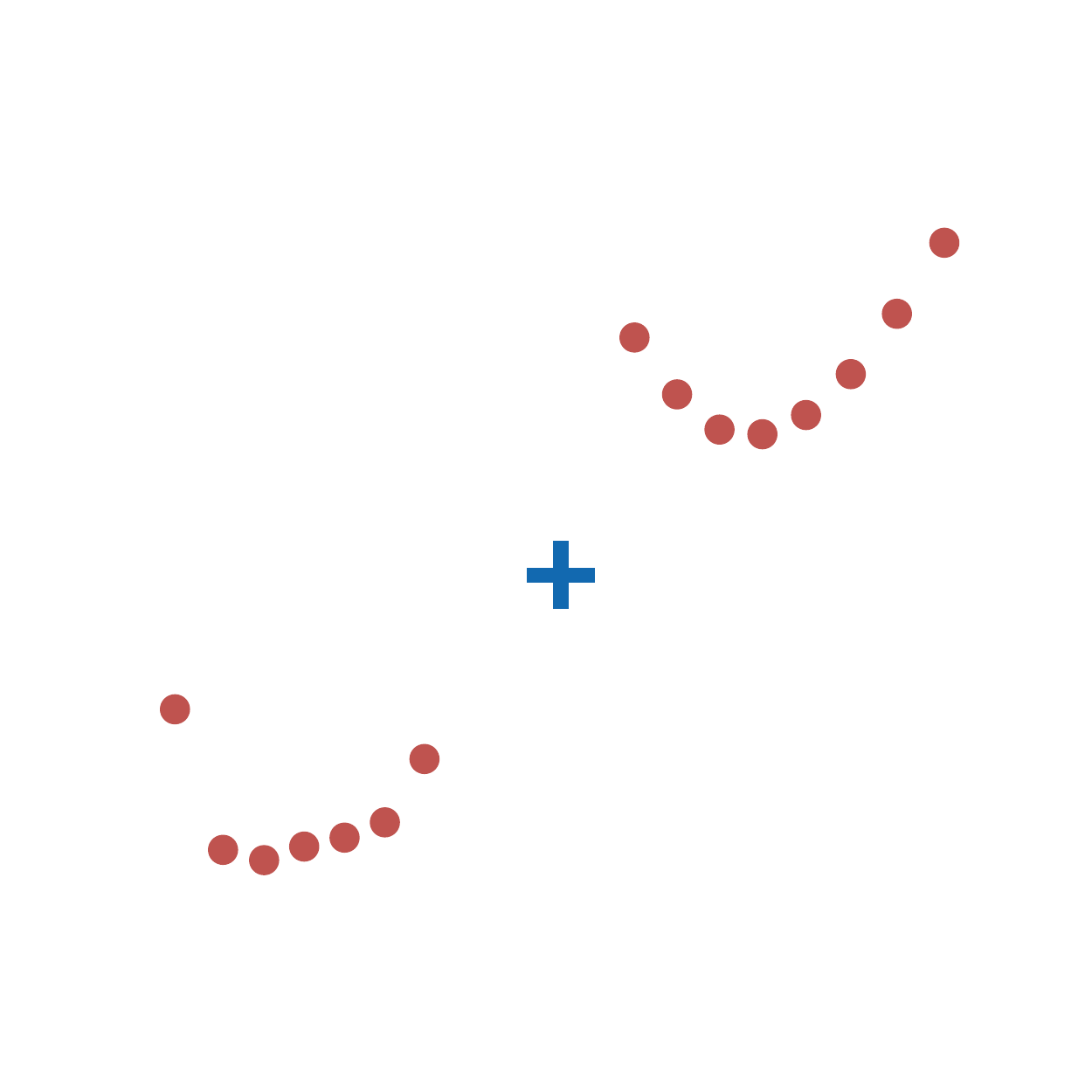}%
    \impair{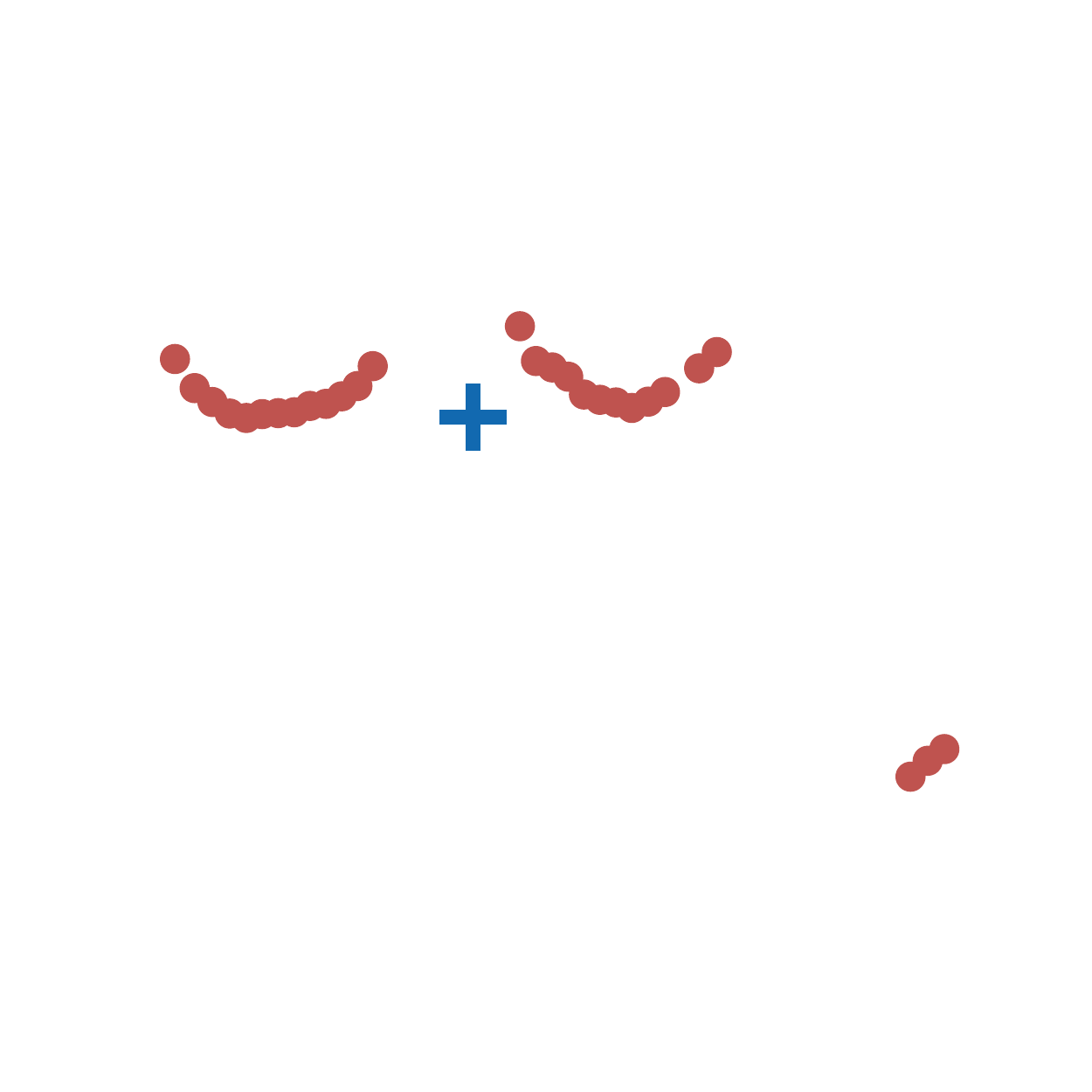}%
    \impair{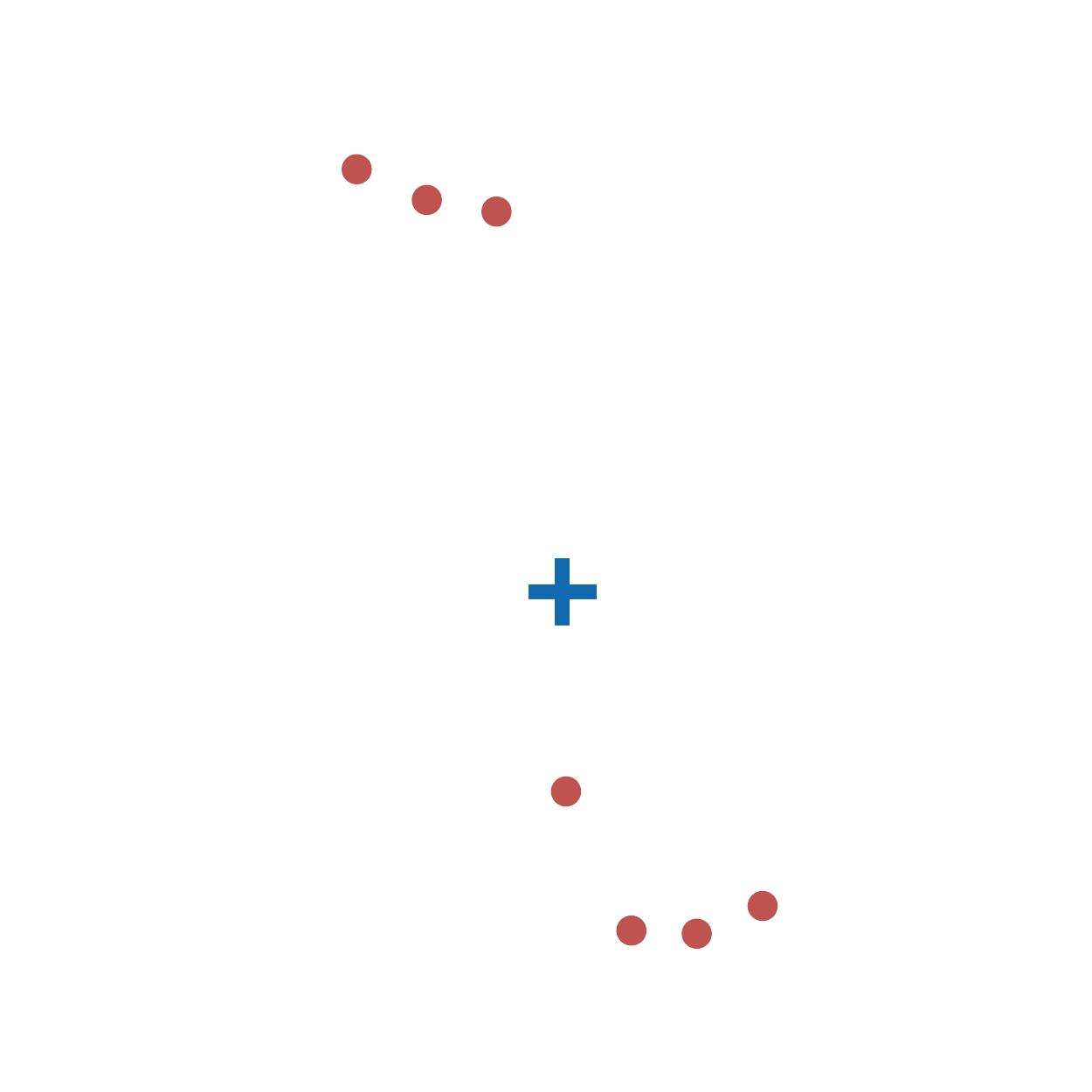}%
    \impair{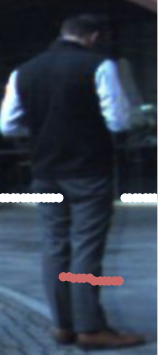}%
    \impair{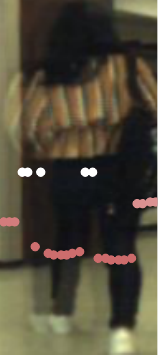}%
    \impairend{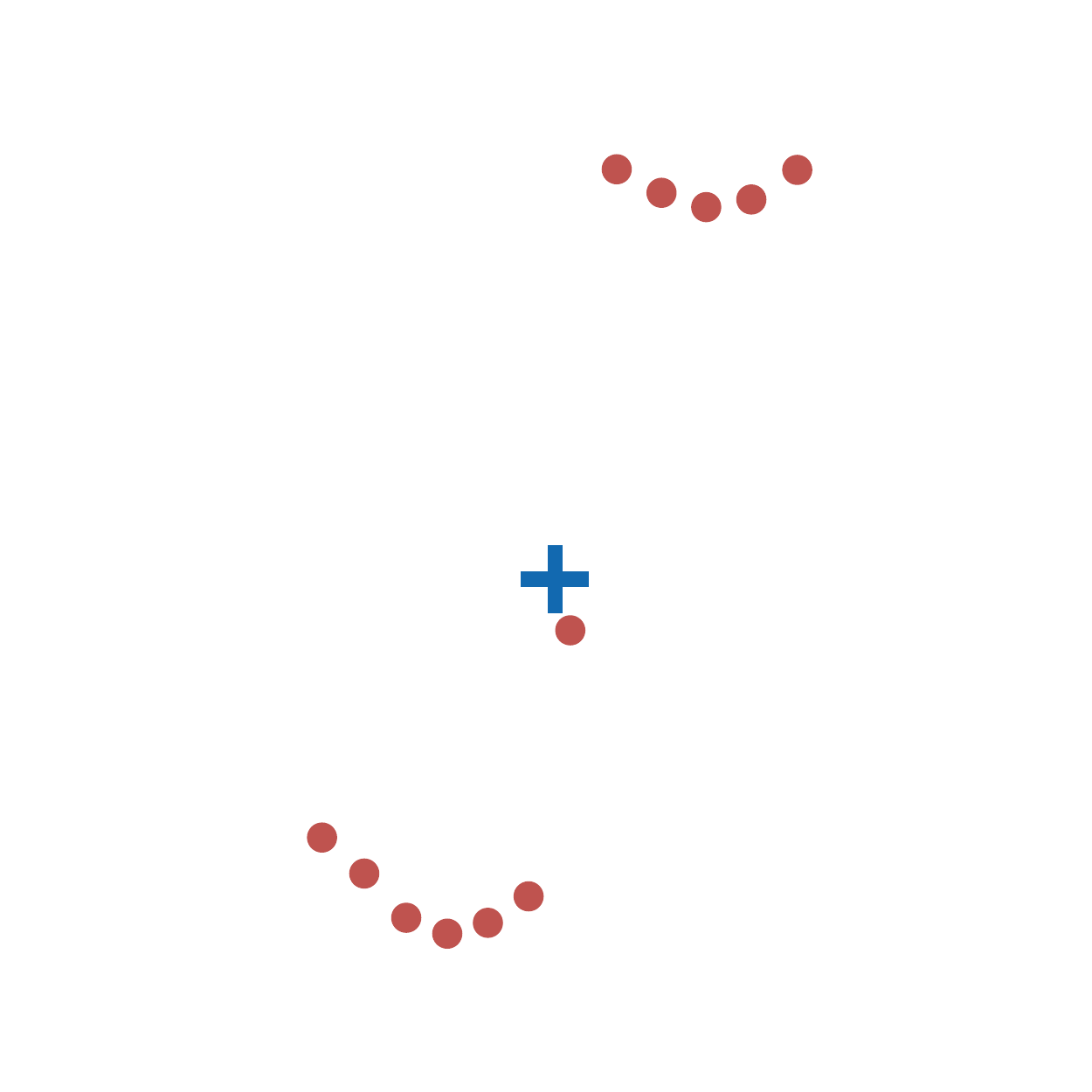}%
    \vspace{2mm}%
    
    \rotatebox[origin=c]{90}{Success}\hspace{2mm}
    \impair{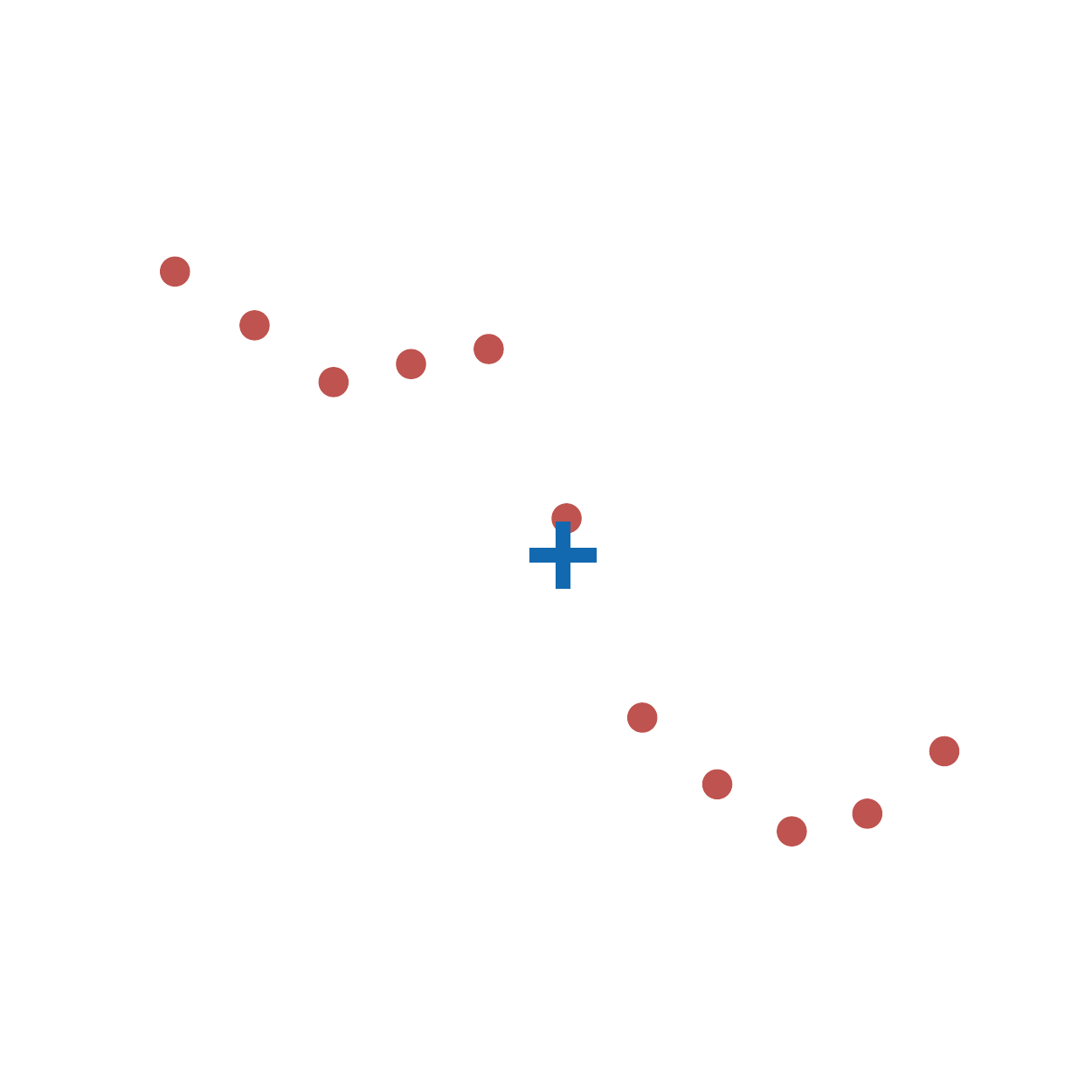}%
    \impair{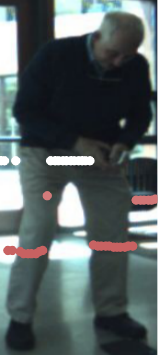}%
    \impair{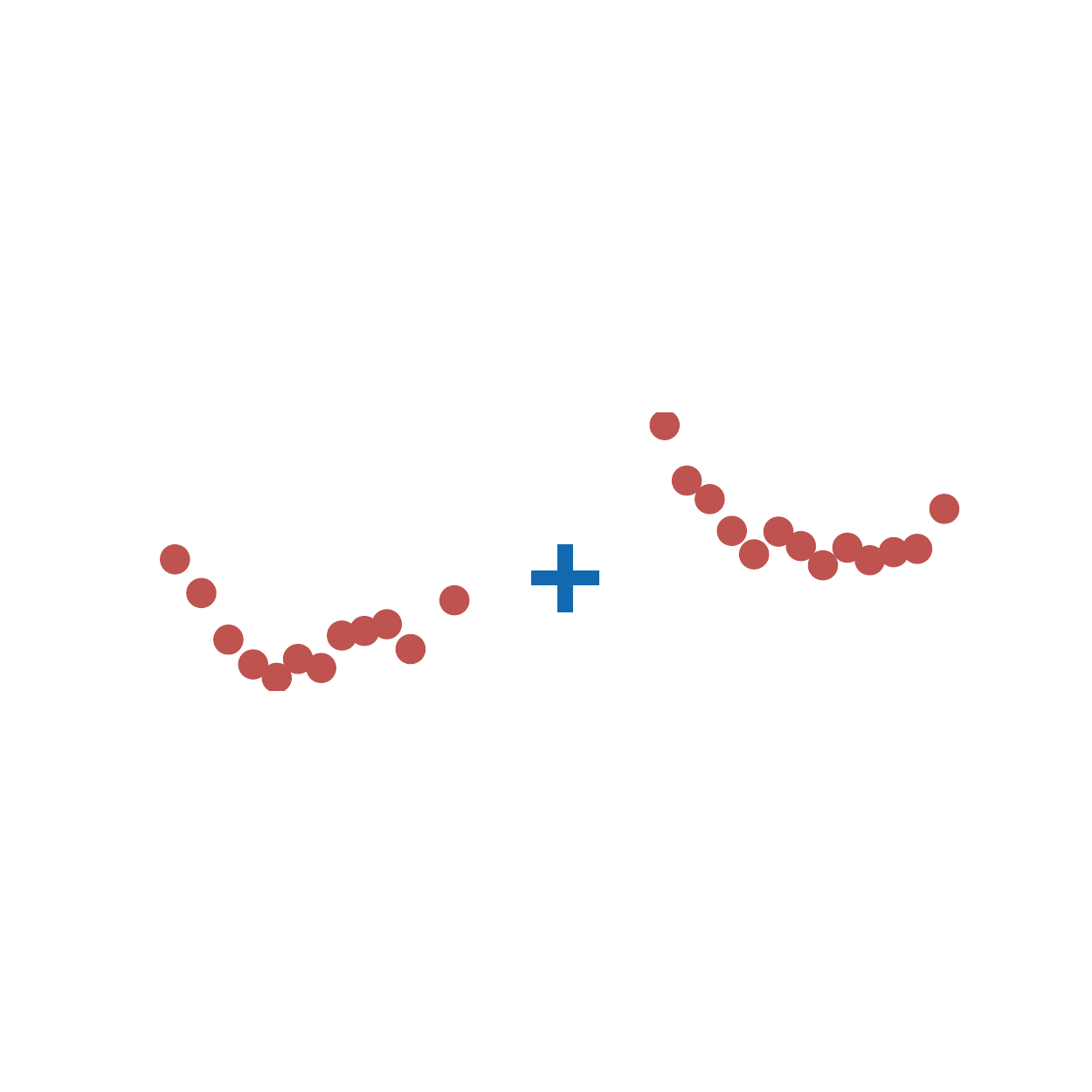}%
    \impair{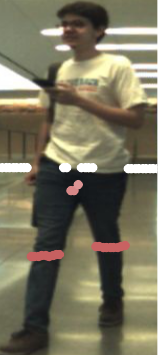}%
    \impair{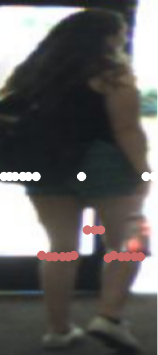}%
    \impairend{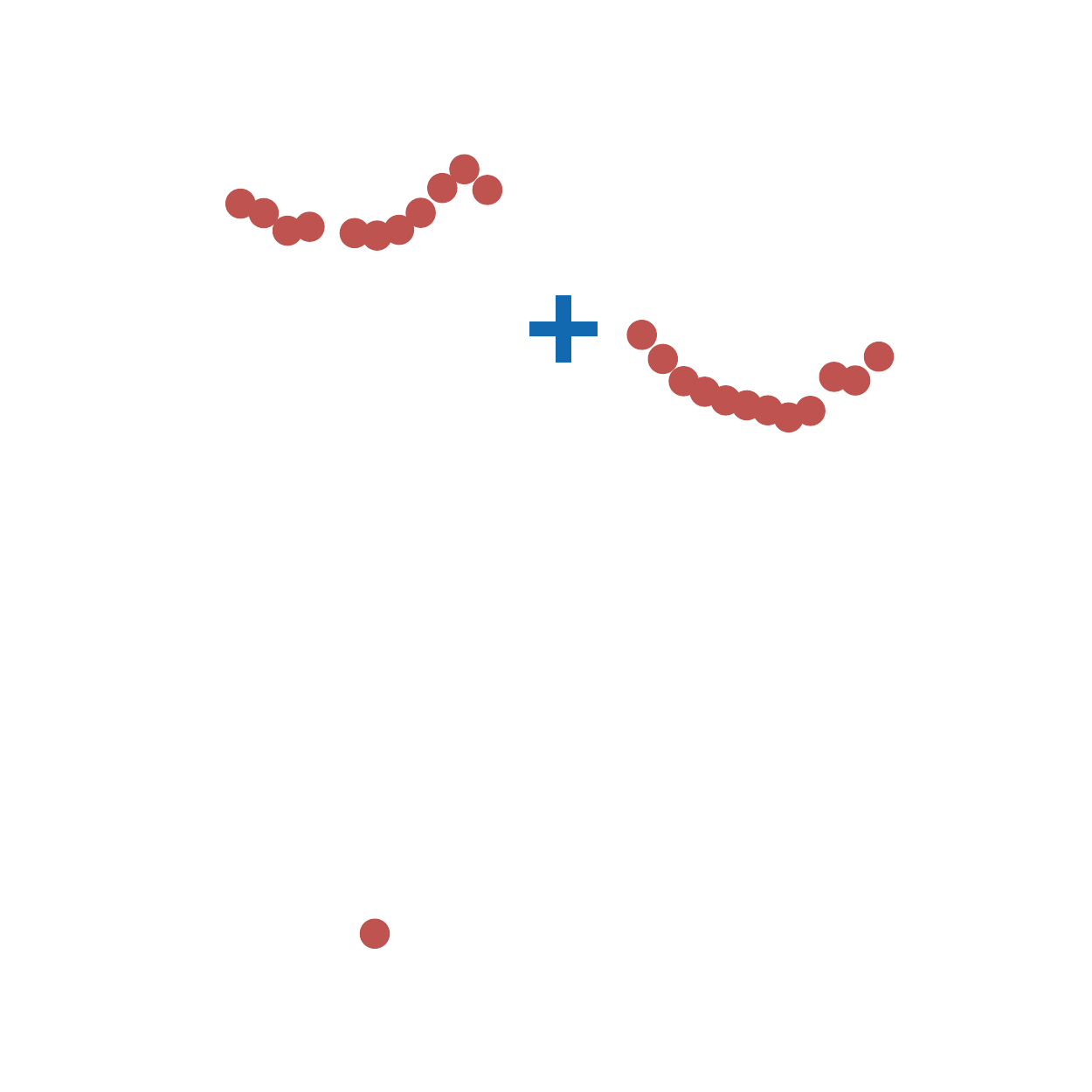}%
    \vspace{2mm}%
    
    \rotatebox[origin=c]{90}{Failure}\hspace{2mm}
    \impair{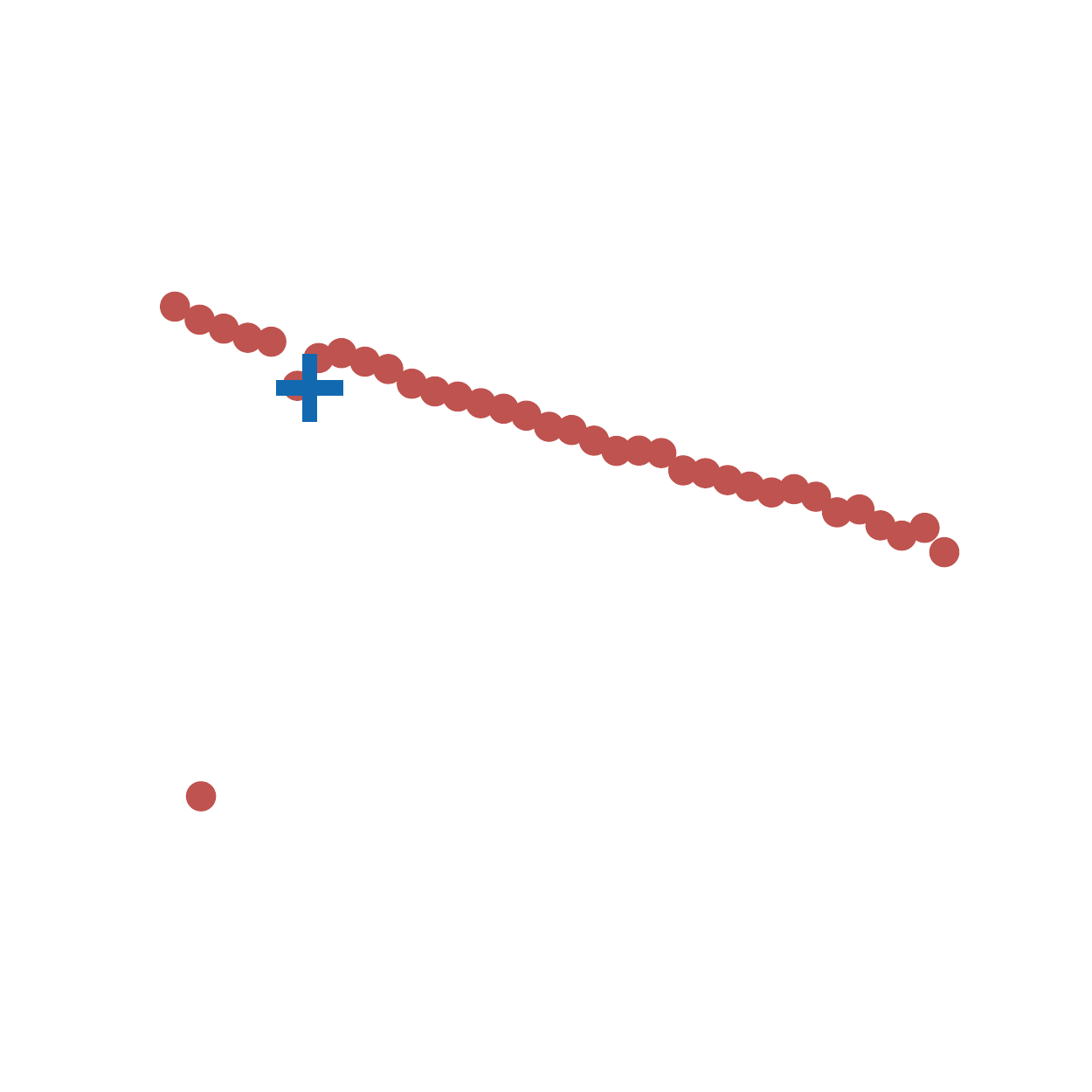}%
    \impair{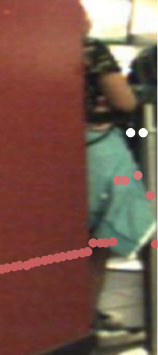}%
    \impair{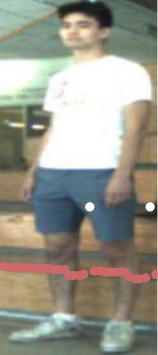}%
    \impair{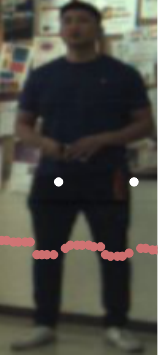}%
    \impair{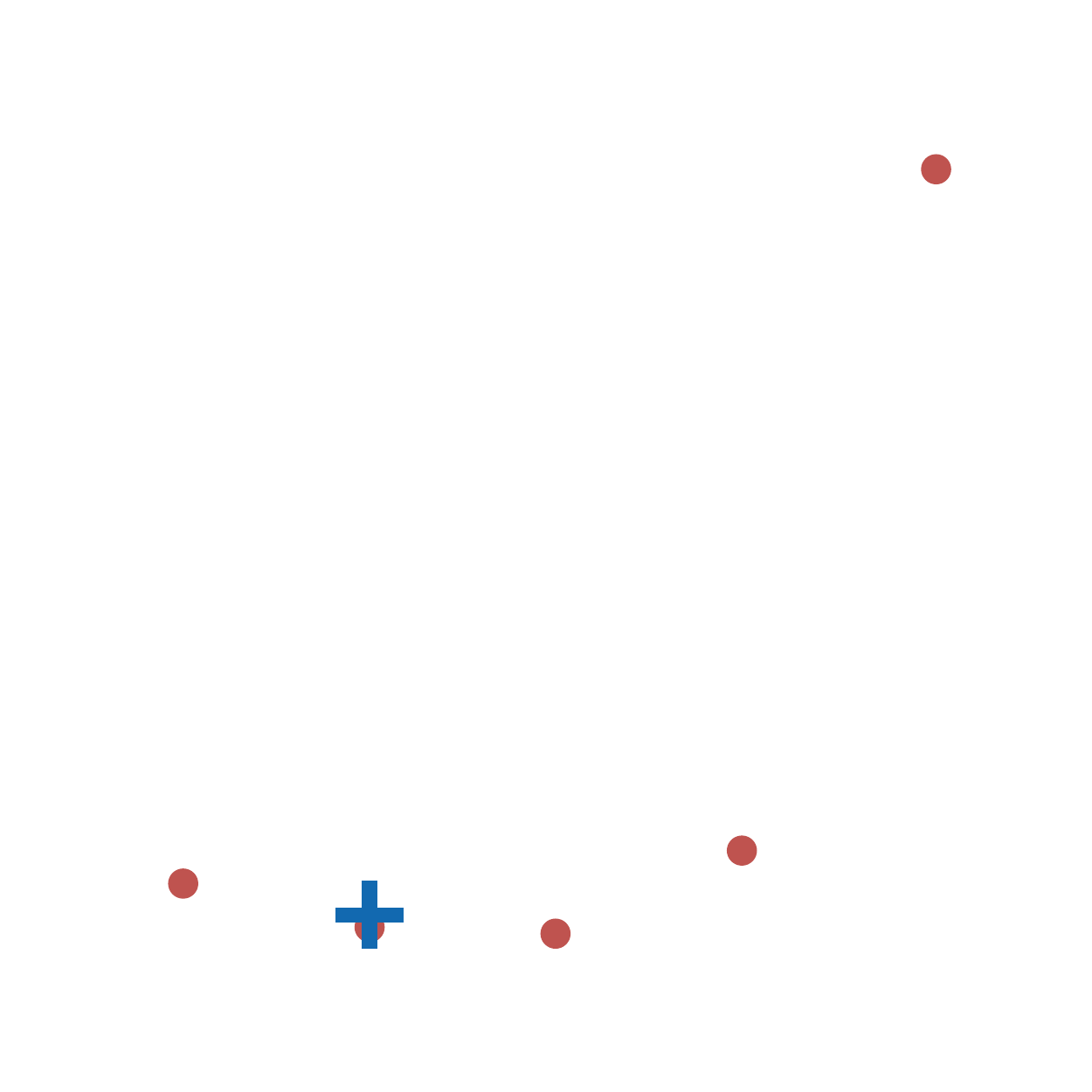}%
    \impairend{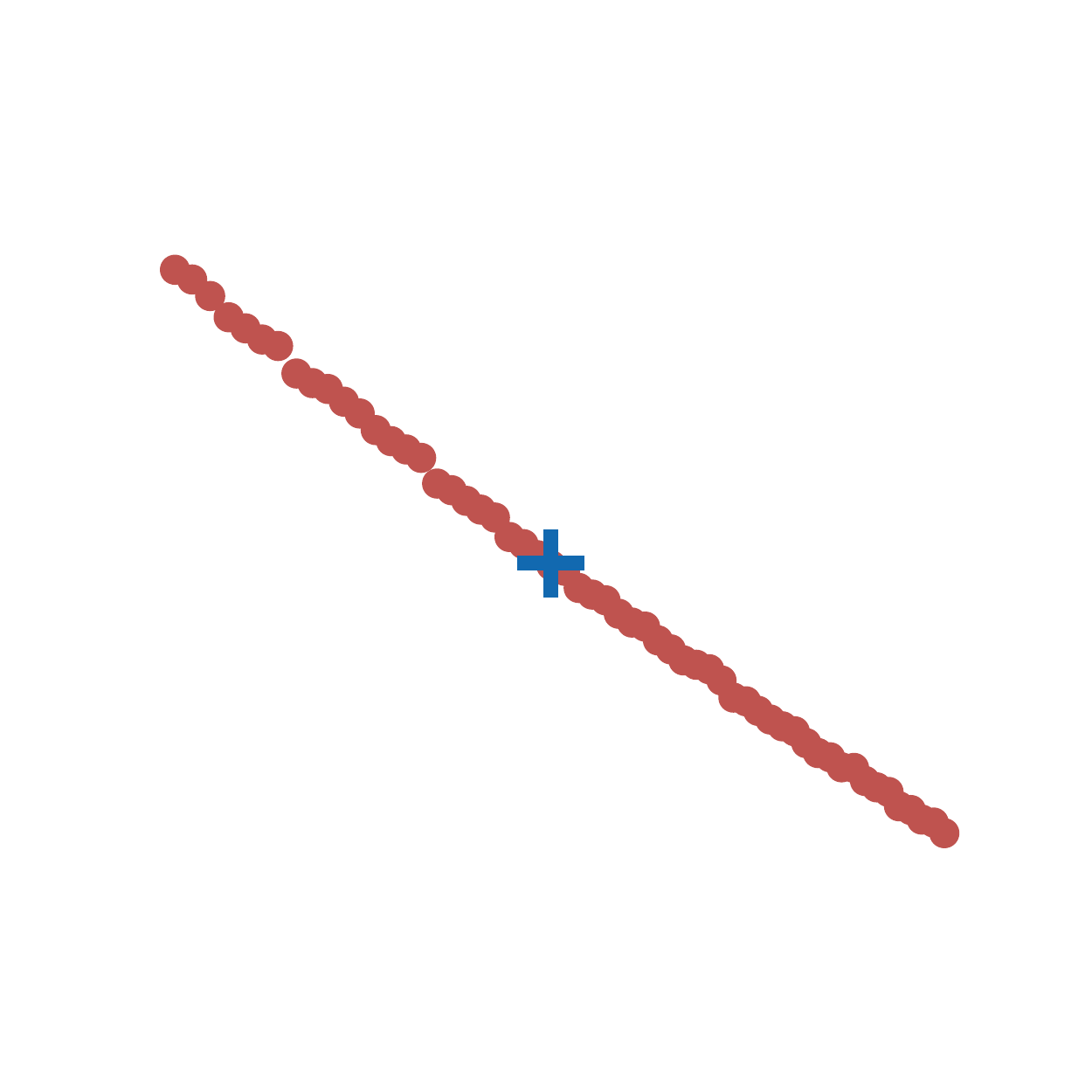}%
    \vspace{2mm}%
    
    \rotatebox[origin=c]{90}{Failure}\hspace{2mm}
    \impair{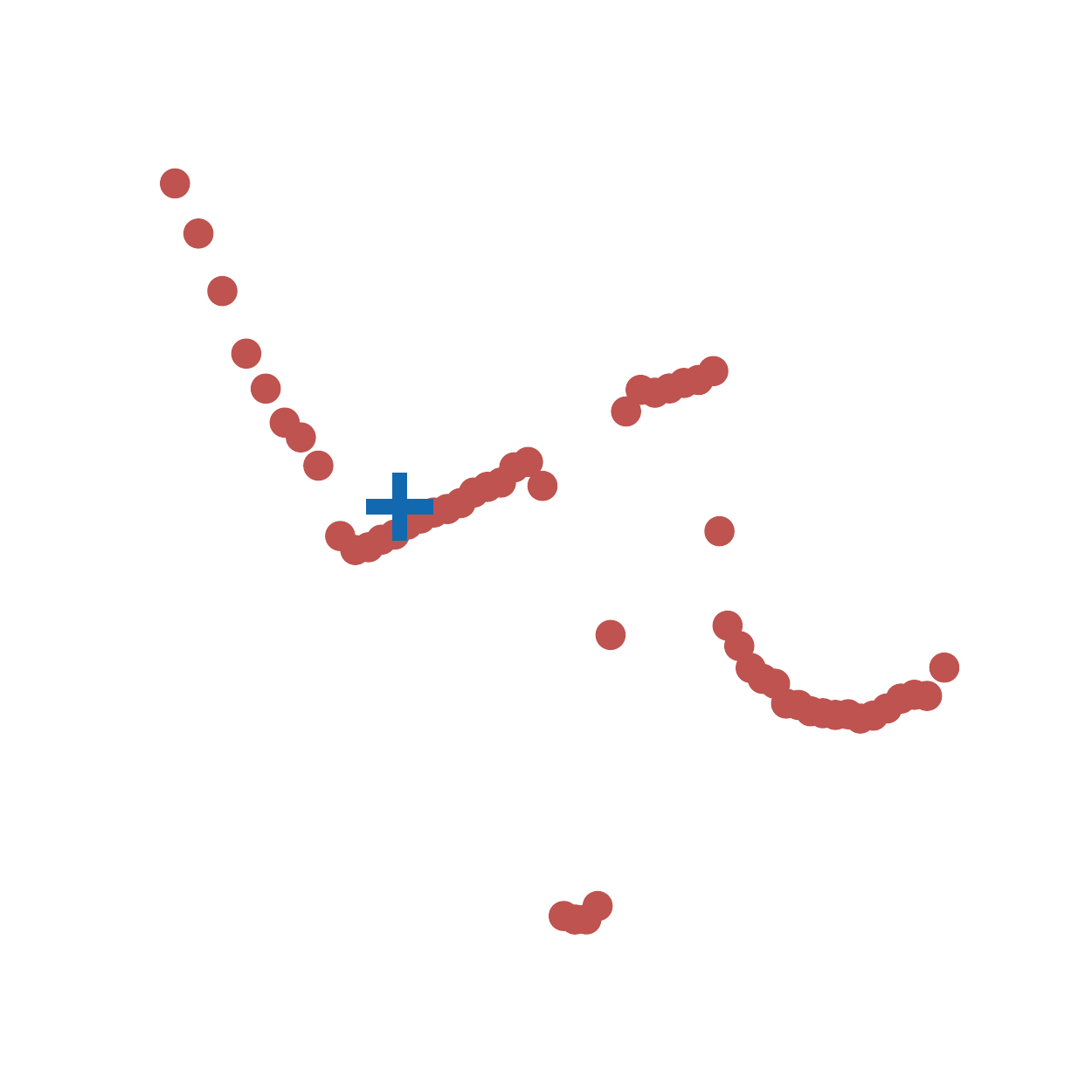}%
    \impair{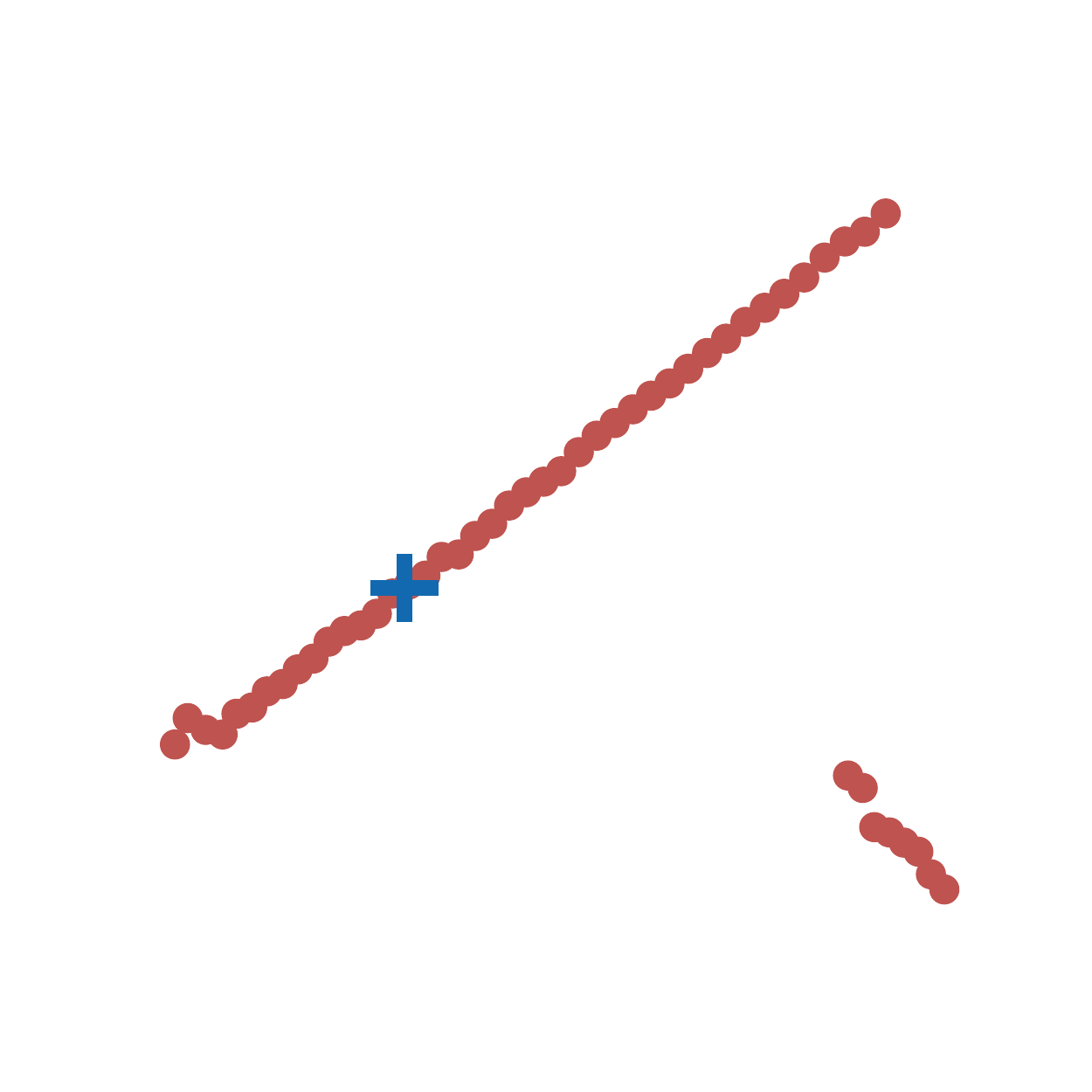}%
    \impair{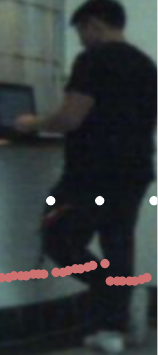}%
    \impair{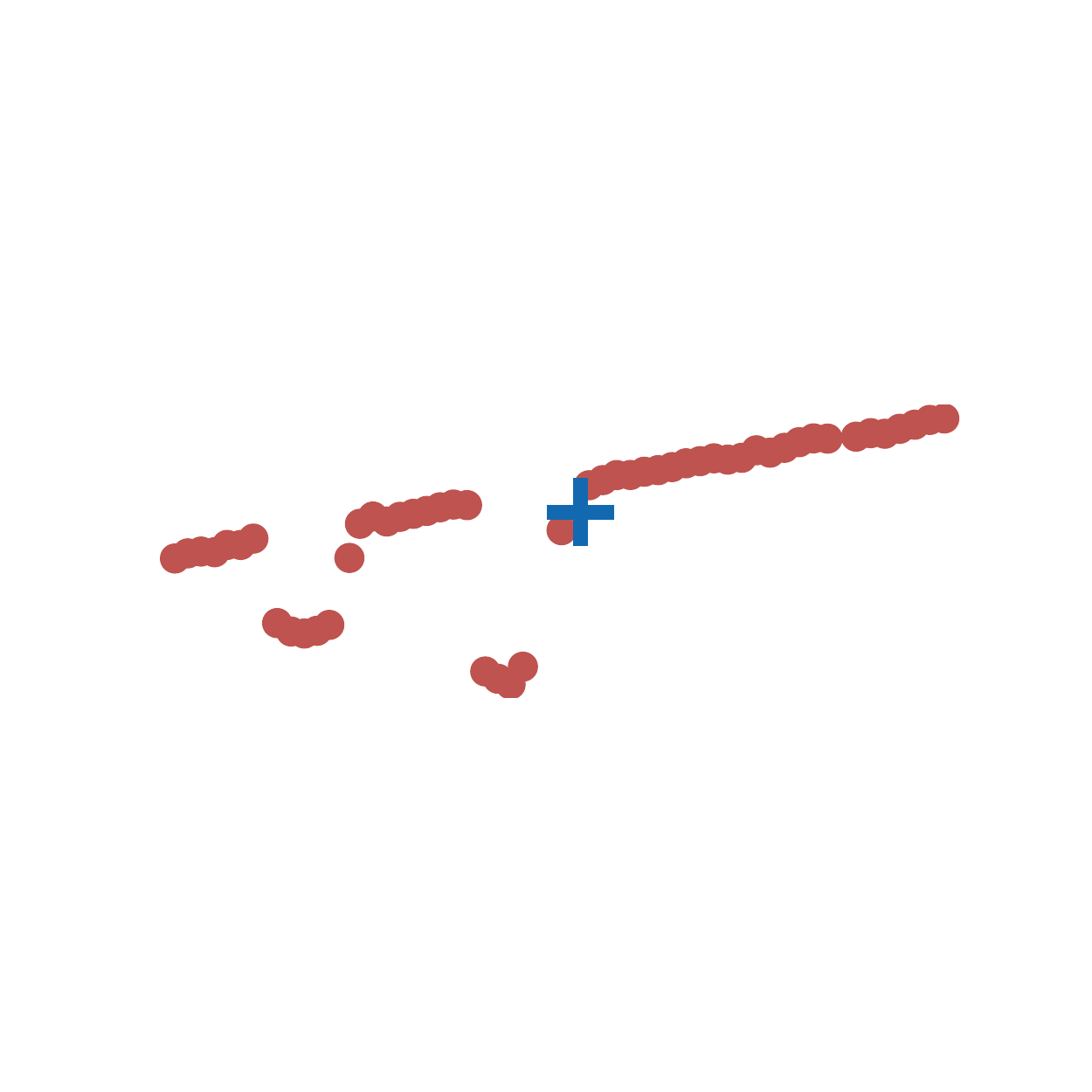}%
    \impair{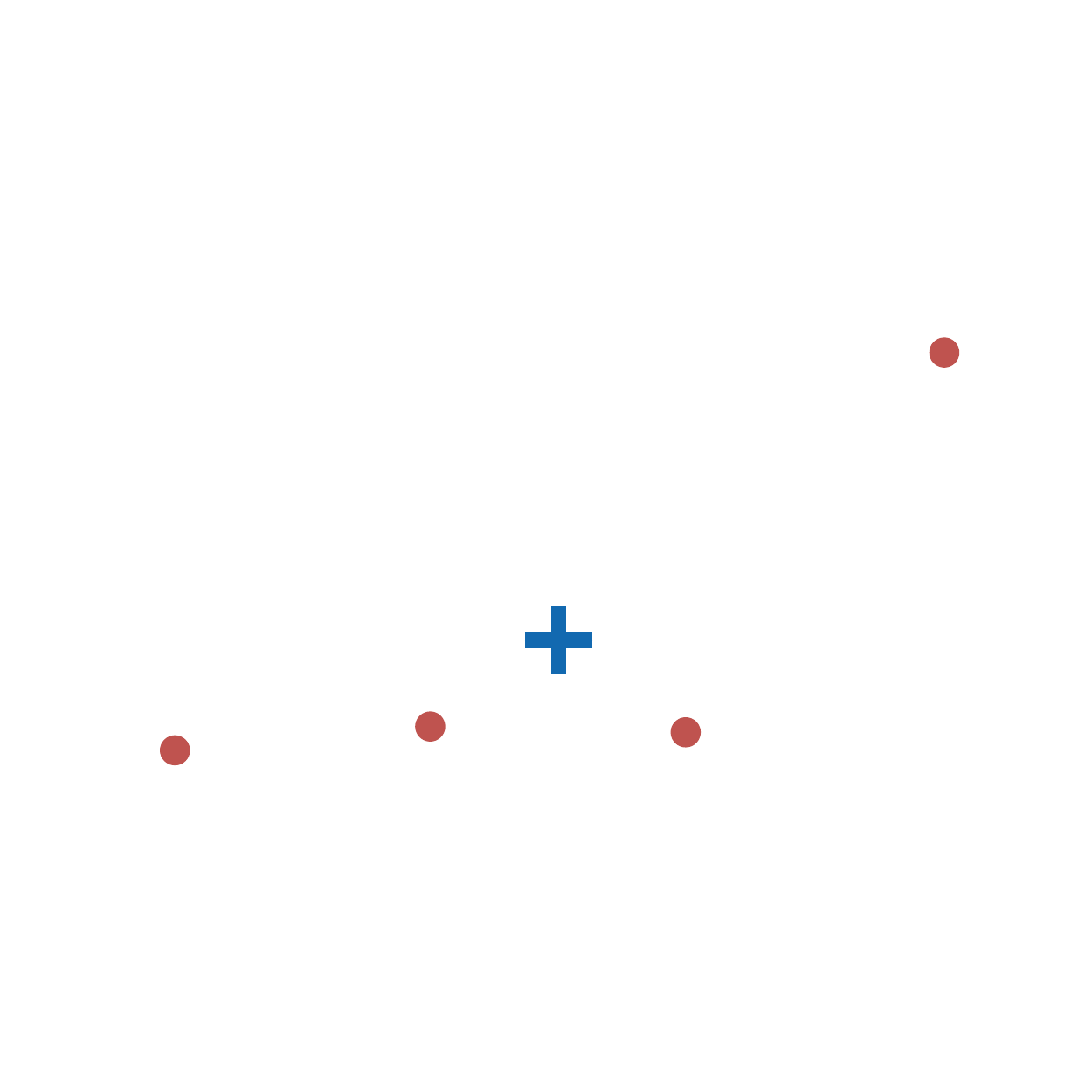}%
    \impairend{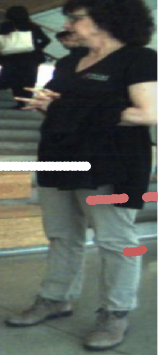}%

    \newcommand\drawcross{%
        \begin{tikzpicture}%
            \draw[color=new_blue_dark,line width=1.5pt]%
                (-3pt,0) -- (3pt,0)%
                (0,-3pt) -- (0,3pt);%
        \end{tikzpicture}%
    }   
    
    \caption{Person detections and the top-down view of the matching pseudo-labels (\raisebox{-0.5mm}{\protect\drawcross{}}) with surrounding LiDAR points (within a 0.5 $m$ radius).
    The LiDAR points are overlaid in the detection images, where the color encodes measured distance with red being closest and white being farthest. 
    The bottom two rows demonstrate some failure cases: heavy occlusion (columns 1-2); background distraction (column 3-4); sparse LiDAR points at far distance (column 5); and faulty calibration or synchronization between sensors (column6).}
    \label{fig:qualitative_results}
\end{figure*}

\section{EVALUATION}
\label{sec:evaluation}

We first conduct experiments to validate the quality of the pseudo-labels and then demonstrate their effectiveness for training person detectors.
We present two case studies: training detectors using pseudo-labels with pre-collected data as well as fine-tuning detectors using dynamically generated pseudo-labels.

\subsection{The JackRabbot Dataset (JRDB)}

Our experiments are conducted on the JackRabbot dataset~\cite{Martin2019arXiv}.
The dataset was collected using a mobile robot, the \textit{JackRabbot}, in both indoor and outdoor environments.
It includes point clouds from 3D LiDARs, annotated with 3D bounding boxes for persons, and RGB camera images, annotated with 2D bounding boxes.
Although not directly annotated, the JackRabbot dataset contains scans from two SICK 2D~LiDARs.
These LiDARs were mounted at the height of the lower legs, facing the front and the back of the robot, respectively.
The dataset features a full 360\degree~scan with 1091 points, generated by combining scans from the two LiDARs, which was used as the LiDAR input in our experiments.

For the evaluation of our pseudo-labels and trained detectors, we convert the annotated 3D bounding boxes into 2D LiDAR annotations by using their center as the ground truth location of persons in the scene.
Since many 3D bounding boxes are occluded in the view of the 2D LiDARs, we only keep annotations that have at least five points within a 0.5~$m$ radius.

The dataset includes person detections from a Faster R-CNN detector~\cite{Ren15NIPS}, which we use as the input to our pseudo-label generation approach. 
However, using the included RGB images other detectors can be applied, allowing us to potentially further improve the label quality.
For generating pseudo-labels, we use $T_{c}=0.75$, $T_{AR}=0.45$, and $T_o=0.4$.

The JRDB does not provide a train-validation split and the test set annotations are not publicly available.
Hence, we employ a custom train-test split.
We split the 27 sequences of the original train set into 17 sequences for training and 10 sequences for testing.
Our train-test split is balanced with respect to person detection difficulty (assessed using a pre-trained DROW3 detector) and scene properties (indoor \textit{vs.} outdoor).
We refer readers to the released code for further details.

\subsection{Pseudo-Label Statistics}
\label{ssec:pseudo_label_quality}

To evaluate the quality of our pseudo-labels, we calculate the true positive and the true negative rate (TPR, TNR) of the classification target generated using pseudo-labels by comparing them against the target generated using ground truth annotations.
More than 90 percent of the training samples are labeled correctly (see~Table\,\ref{table:pl_pr_tnr}), indicating the validity of the pseudo-labels.
Qualitative results of pseudo-labels are shown in~Fig.\,\ref{fig:qualitative_results}.
In most cases, locations of persons are successfully estimated in both indoor and outdoor environments, with people at different distances or having different poses.
Common failure cases are based on occluding objects, persons merging with the background, sparse LiDAR measurements at high distances, or a faulty calibration between sensors.
These cases result in noisy labels, which we deal with by using a robust training loss.


\begin{table}[b]
\centering
\setlength{\tabcolsep}{2.0pt}
\begin{tabularx}{0.8\columnwidth}{ l ccc YY }
\toprule
Bounding Boxes      &&&& TPR & TNR  \\ 
\midrule
Faster R-CNN Detections   &&&& 91.6 & 99.4\\
2D Annotations      &&&& 90.6 & 99.1\\
\bottomrule
\end{tabularx}
\caption{Accuracy of pseudo-labels on the training split.}
\label{table:pl_pr_tnr}
\end{table}


To ablate the effect of uncertainty in the image-based person detector, we generate pseudo-labels using the annotated 2D bounding boxes.
The TPR and TNR of the pseudo-labels generated using detections are higher than those generated using annotations, showing that our proposed method is robust against uncertainty in bounding boxes.
Due to the sparsity of LiDAR points, it is difficult to generate pseudo-labels for far away persons contained in 2D annotations.
When using a detector, these persons are often missed or detected with low confidence, thus not leading to a (potentially false) pseudo-label.

\begin{figure}
    \includegraphics[width=\linewidth]{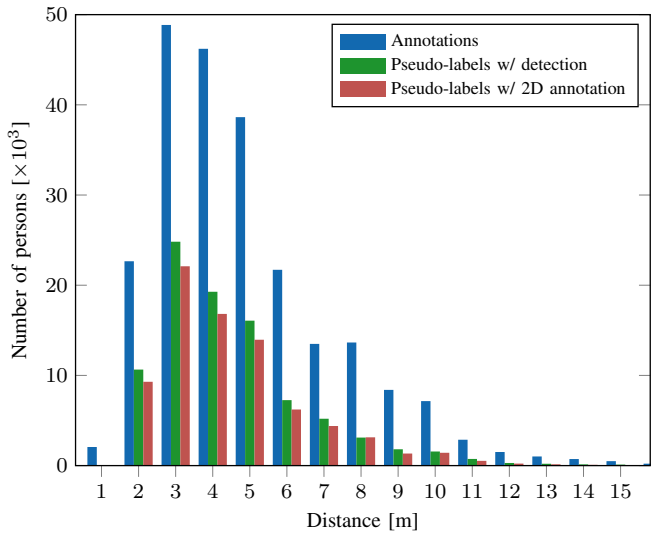}%
    \caption{
    Distance distribution of pseudo-labels and annotations.
    Pseudo-labels have a similar distance distribution to that of the annotations and do not introduce a distance bias.
    }%
    \label{fig:pl_dist}%
\end{figure}

We also analyze the distance distribution of our pseudo-labels.
As Fig.\,\ref{fig:pl_dist} shows, it is similar to that of the annotations.
Pseudo-labels do not introduce any distance bias which may wrongly emphasize samples within a certain range.
Due to the high mounting of cameras in the JackRabbot dataset, the legs of persons close to the robot are outside the camera's field of view.
Thus, no pseudo-label can be generated within 1~$m$ distance.


\subsection{Training with Pseudo-Labels}
\label{sec:offline_training}

We examine the performance of detectors trained using either pseudo-labels or ground truth annotations.
We train the DROW3 detector for 40 epochs, using a batch size of 8 scans and the DR-SPAAM detector for 20 epochs, using a batch size of 4 scans,
We use the Adam optimizer~\cite{Kingma15ICLR}, and a learning rate of $10^{-3}$.
Starting from 10 epochs for DROW3 and 5 epochs for DR-SPAAM, we exponentially decay the learning rate until it reaches $10^{-6}$ at the end of training.
Implementation and hyper-parameters are taken from~\cite{Jia20arXiv} for both detectors, with the only exception being that we enlarge the spatial window size for DR-SPAAM to 17 points (from the original 11 points).
This enlarged window matches the effective opening angles, since the scans in JRDB have a higher angular resolution.
Due to GPU memory constraints, we randomly crop the scans to 1000 points for DR-SPAAM at training time.

We report the average precision (AP$_{\text{0.3}}$ and AP$_{\text{0.5}}$) on the test split (see~Table\,\ref{table:result_train}).
A detection is considered positive if there is a ground truth within 0.3~$m$ or 0.5~$m$, and a ground truth can only be matched with a single detection.
As baselines, we evaluate the released DROW3 and DR-SPAAM model from~\cite{Jia20arXiv}, which is trained on the DROW dataset.
These pre-trained networks have significantly lower AP compared to the ones trained on JRDB, confirming our initial speculation that networks trained on a single dataset may not generalize well to new environments or different LiDAR models.
The pre-trained DR-SPAAM, despite having a higher score on the DROW dataset, performed worse than DROW3, showing the effect of overfitting.
Networks trained using pseudo-labels, benefiting from a smaller domain gap to the test data, outperform the pre-trained networks, proving the validity of our approach.
Starting from a pre-trained model improves the detector performances for pseudo-labels, but gives no clear improvement for 3D annotations.


The performance gap between pseudo-labels and annotations could be caused by two factors: label noise and less training samples.
To study the effect of label noise, we additionally train networks with pseudo-labels, while removing falsely labeled points and correcting the regression target using ground truth annotations (see~Table\,\ref{table:result_train_pl}).
Both false positives and false negatives reduce the detector performance, with the later having a strong influence on DROW3.
Correcting the regression target improves AP$_\text{0.3}$ for both networks, especially for the more powerful DR-SPAAM.
Cleaning pseudo-labels increases detector performance significantly, showing that label noise is the more dominant cause to the performance gap between pseudo-labels and annotations.
Detectors trained using clean pseudo-labels trail the ones trained using annotations by around 2 percent AP, due to reduced amounts of training samples.
Fine-tuning with a pre-trained model further narrows this performance gap.

\begin{table}
\centering
\setlength{\tabcolsep}{2.0pt}
\begin{tabularx}{\linewidth}{l cc YYccYY }
\toprule 
&&& \multicolumn{2}{c}{DROW3} &&& \multicolumn{2}{c}{DR-SPAAM} \\
\cmidrule{4-5} \cmidrule{8-9}
Supervision &&& AP$_{\text{0.3}}$ & AP$_{\text{0.5}}$ & & & AP$_{\text{0.3}}$ & AP$_{\text{0.5}}$  \\ 
\midrule
Pre-trained from \cite{Jia20arXiv} &&& 65.5 & 70.8 &&& 62.5 & 68.3 \\
Pseudo-labels &&& 68.5 & 77.3 &&& 66.9 & 75.5 \\
\quad + fine-tuning from \cite{Jia20arXiv} &&& 69.0 & 77.8 &&& 69.2 & 76.7 \\
3D Annotation &&& 76.2 & \textbf{82.9} &&& 78.5 & \textbf{84.9} \\
\quad + fine-tuning from \cite{Jia20arXiv} &&& \textbf{76.4} & 82.5 &&& \textbf{78.6} & 83.8 \\
\bottomrule
\end{tabularx}
\caption{Performance of DROW3 and DR-SPAAM trained using different supervision.}
\label{table:result_train}
\end{table}

\begin{table}
\centering
\setlength{\tabcolsep}{2.0pt}
\begin{tabularx}{\linewidth}{l c YYccYY }
\toprule 
&& \multicolumn{2}{c}{DROW3} && \multicolumn{2}{c}{DR-SPAAM} \\
\cmidrule{3-4} \cmidrule{6-7}
Supervision && AP$_{\text{0.3}}$ & AP$_{\text{0.5}}$ && AP$_{\text{0.3}}$ & AP$_{\text{0.5}}$  \\ 
\midrule
Pseudo-labels && 68.5 & 77.3 && 66.9 & 75.5 \\
$\ldots$ (remove FP) && 71.6 & 79.0 && 71.1 & 77.8 \\
$\ldots$ (remove FN) && 68.4 & 77.8 && 70.3 & 78.6 \\
$\ldots$ (remove FP \& FN) && 73.1 & 80.8 && 72.2 & 78.8 \\
$\ldots$ (remove FP \& FN, correct reg.) && 74.6 & 80.5 && \textbf{76.5} & \textbf{81.5} \\
\quad + fine-tuning from \cite{Jia20arXiv} && \textbf{74.7} & \textbf{81.2} && 76.2 & \textbf{81.5} \\
\bottomrule
\end{tabularx}
\caption{Performance of DROW3 and DR-SPAAM trained with different variants of pseudo-labels.}
\label{table:result_train_pl}
\end{table}

\begin{table}
\centering
\setlength{\tabcolsep}{2.0pt}
\begin{tabularx}{\linewidth}{l cc YYccYY }
\toprule 
&&& \multicolumn{2}{c}{DROW3} &&& \multicolumn{2}{c}{DR-SPAAM} \\
\cmidrule{4-5} \cmidrule{8-9}
Training scheme &&& AP$_{\text{0.3}}$ & AP$_{\text{0.5}}$ & & & AP$_{\text{0.3}}$ & AP$_{\text{0.5}}$  \\ 
\midrule
Cross-entropy loss &&& 68.5 & 77.3 &&& 66.9 & 75.5 \\
\quad + mixup regularization &&& 69.5 & 78.0 &&& 65.8 & 74.0 \\[0.1cm]
Partially Huberized cross-entropy loss &&& \textbf{71.4} & \textbf{79.0} &&& 69.4 & 76.4 \\
\quad + mixup regularization &&& 71.1 & 78.5 &&& \textbf{70.0} & \textbf{78.3} \\
\bottomrule
\end{tabularx}
\caption{Performance of DROW3/DR-SPAAM trained with pseudo-labels and different robust training methods.}
\label{table:result_robust}
\end{table}

To mitigate the problem caused by labeling noise, we experiment with two robust training methods: the \textit{partially Huberized cross-entropy loss}, and the \textit{mixup} regularization (see~Table\,\ref{table:result_robust}).
Both methods improve the network performance, with the exception of applying \textit{mixup} alone on DR-SPAAM (potentially due to suboptimal hyper-parameters).
Combined with robust training methods, detectors trained using pseudo-labels outperform the pre-trained detectors by a large margin, and reach a performance close to training using annotations, without using any labeled data.
Pseudo-labels provide an effective way to adjust person detectors to new environments or LiDAR models.


\subsection{Online Fine-Tuning with Pseudo-Labels}

In practice, it is desirable to use a detector that can dynamically fine-tune itself in an online fashion during deployment.
To study the network performance undergoing such fine-tuning, we take a DROW3 detector, pre-trained on DROW dataset~\cite{Jia20arXiv}, and fine-tune it on the JRDB train split for one epoch, with the partially Huberized cross-entropy loss.
We use a learning rate of $5\times10^{-5}$ and a batch size of 8 scans.
In the first set of experiments, we shuffle data only within each sequence (the whole train split is composed of 17 sequences) and pass the in-sequence shuffled data to the network.
This mimics the situation where a mobile robot enters into a new environment, curates a small amount of data, and fine-tunes itself.
In the second set of experiments, we shuffle data within the whole training split, giving more diversity in each batch.

\begin{figure}
    \newcommand\drawdashedline{%
        \begin{tikzpicture}%
            \draw[gray, dashed, line width=0.75pt]%
                (-7.5pt,0) -- (7.5pt,0);%
        \end{tikzpicture}%
    }   
    \centering
    \includegraphics[width=\linewidth]{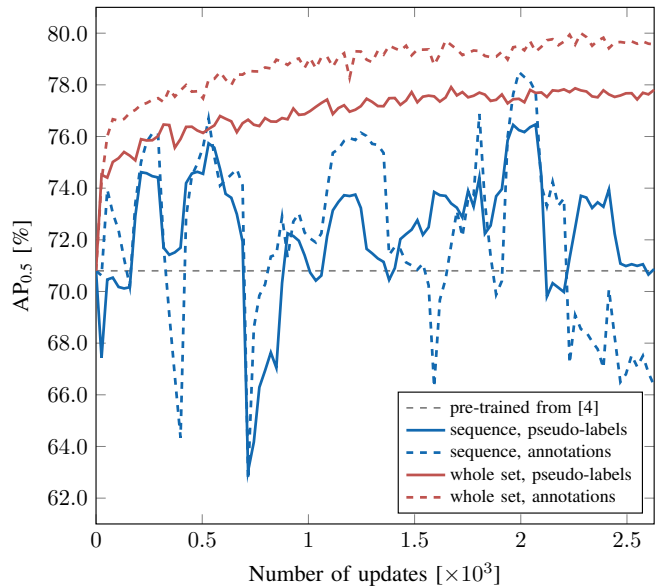}
    \caption{
    Performance at different steps of fine-tuning a pre-trained DROW3 detector with partially Huberized cross-entropy loss.
    Data is shuffled either on the whole training set or within each sequence, mimicking different amounts of curated data.
    With diverse batches, the detector performance improves significantly even with only a small number of updates.
    Fine-tuning longer using the default training schedule yields 79.2 and 82.3 percent AP$_\text{0.5}$ for pseudo-labels and annotations, respectively.  
    }

    \label{fig:online_training}
\end{figure}

The detector performance at different stages of fine-tuning is shown in~Fig.\,\ref{fig:online_training}.
When the data is shuffled within the whole training split, the network performance increases significantly, from the pre-trained 70.8 percent AP$_{\text{0.5}}$ to more than 74 percent, using less than one hundred updates.
This fast performance increase implies that, even in applications with insufficient computation for a full training, it is still possible to adapt the detector and improve its performance, by running a small number of updates using pseudo-labels.
However, having curated training samples with enough diversity is a key prerequisite, as fluctuating performance is observed when the data was shuffled only within each sequence.
Although for most of the time the detector benefits from fine-tuning (having better performance than that of a pre-trained detector), there are adversarial samples that cause dramatic performance reductions.
The same fluctuating behavior exists for fine-tuning using annotations, showing that it is an innate problem of network training, rather than caused by pseudo-labels.
Curriculum learning methods~\cite{Bengio09ICML,Jiang18ICML} may help mitigating this problem, and thus relaxing the requirement of data collection. 
They are interesting directions to be explored in future research.


\section{CONCLUSION}

In this paper, we proposed a method to automatically generate pseudo-labels for training 2D LiDAR-based person detectors, using bounding boxes generated from an image-based person detector with a calibrated camera.
We analyzed the quality of pseudo-labels by comparing them against ground truth annotations and proved their validity.
Experiments were conducted to train or fine-tune DROW3 and DR-SPAAM detectors using pseudo-labels, and these self-supervised detectors outperformed the detectors trained on annotations using a different dataset.
Even stronger detectors were obtained by combining pseudo-labels with robust training techniques.

Our method provides an effective way to bridge the domain gap between data encountered during training and during deployment.
With our method, a mobile robot equipped with a 2D LiDAR-based person detector can fine-tune the detector during deployment, improving its performance with no additional labeling effort.
With the released code, we expect our method will be useful for many robotic applications.


\textbf{Acknowledgements:}

We thank Hamid Rezatofighi, JunYoung Gwak, and Mihir Patel for their help with the JackRabbot dataset.
This project was funded by the EU H2020 project "CROWDBOT" (779942). 
Most experiments were performed on the RWTH Aachen University CLAIX 2018 GPU Cluster (rwth0485).


\newpage
\balance
\bibliographystyle{ieeetr}
\bibliography{abbrev,mybib}



\end{document}